\documentclass[conference]{IEEEtran}
\usepackage{url}
\usepackage{color}
\usepackage{algorithm,algorithmicx}
\usepackage{empheq,hhline}
\usepackage[noend]{algpseudocode}

\usepackage{amsmath,amsthm,amssymb,url,graphicx,rotating,ifthen,epsfig,array,caption,color}
\usepackage{todonotes}
\usepackage{graphicx}
\usepackage{caption}
\usepackage{subcaption}
\usepackage{balance}
\usepackage{multirow}
\usepackage{enumitem}
\def\x{\mathbf{x}}

	\def\G{{\mathcal N}}
\def\P{{\cal P}}
\def\R{{\mathbb R}}

\ifCLASSINFOpdf
\else
\fi
\begin{document}
%
\title{Efficient Noisy Optimisation with the \\
Sliding Window Compact Genetic Algorithm
}
%
%
%
\def\nouse{
\author{
Simon~M.~Lucas,~\IEEEmembership{Senior Member,~IEEE,}
Jialin~Liu,~\IEEEmembership{Member,~IEEE,}
Diego~Perez-Liebana,~\IEEEmembership{Member,~IEEE}

\thanks{S. M. Lucas, J. Liu, D. Perez-Liebana are with the Department of Computer Science and Electronic Engineering, University of Essex, Colchester CO4 3SQ, UK
(e-mail: sml@essex.ac.uk; jialin.liu@essex.ac.uk; dperez@essex.ac.uk).}
}
}
\author{
\IEEEauthorblockN{Simon M. Lucas}
\IEEEauthorblockA{Queen Mary University of London\\
London E1 4NS, United Kingdom\\
\url{simon.lucas@qmul.ac.uk}}
\and
\IEEEauthorblockN{Jialin Liu}
\IEEEauthorblockA{Queen Mary University of London\\
London E1 4NS, United Kingdom\\
\url{jialin.liu@qmul.ac.uk}}
\and
\IEEEauthorblockN{Diego P\'erez-Li\'ebana}
\IEEEauthorblockA{University of Essex\\
Colchester CO4 3SQ, United Kingdom\\
\url{dperez@essex.ac.uk}}
}

\maketitle



\begin{abstract}
The compact genetic algorithm is an Estimation of Distribution Algorithm for binary optimisation problems. Unlike  the standard Genetic Algorithm, no cross-over or mutation is involved.  Instead, the compact Genetic Algorithm uses a virtual population represented as a probability distribution over the set of binary strings. At each optimisation iteration, exactly two individuals are generated by sampling from the distribution, and compared exactly once to determine a winner and a loser.  The probability distribution is then adjusted to increase the likelihood 
of generating individuals similar to the winner.

This paper introduces two straightforward variations of the compact Genetic Algorithm, each of which lead to a significant improvement in performance. The main idea is to make better use of each fitness evaluation, by ensuring that each evaluated individual is used in multiple win/loss
comparisons. The first variation is to sample $n>2$ individuals at each iteration to make $n(n-1)/2$
comparisons.  The second variation only samples one individual at each iteration but keeps a sliding
history window of previous individuals to compare with.  We evaluate methods on two noisy test problems and show that in each case they significantly outperform the compact Genetic Algorithm, while maintaining the simplicity of the algorithm.




\end{abstract}

\begin{IEEEkeywords}
Compact genetic algorithm, evolutionary algorithm, estimation of distribution, sliding window, binary optimisation, discrete optimisation.
\end{IEEEkeywords}

%
\IEEEpeerreviewmaketitle

\section{Introduction}

In many optimisation applications there is a need to deal with noisy evaluation functions, and also to make best possible use of a limited evaluation budget. Traditional, population-based evolutionary algorithms most commonly cope with noise by re-evaluating each evaluated individual several times and/or increasing the size of the population. For a recent survey refer to Rakshit et al~\cite{rakshit2016noisy}.

The motivation behind our work is to develop powerful optimisation algorithms that are well-suited to applications in Game AI.  The applications include rolling-horizon planning algorithms used to control non-player characters
or bots, and to automatic game design or automatic game tuning.  Both of these applications have many forms of uncertainty which introduce significant noise into the fitness evaluation function. Furthermore, they each operate
under a limited time budget, so there is a strong need for algorithms that make the best possible use of a limited number of fitness (objective function) evaluations.

The algorithms developed in this paper are already showing promise at the initial testing phase on exactly these types of problems, including General Video Game AI \cite{perez20162014,liu2017evolving}, but in this paper we focus on describing the algorithms and providing results on some simple benchmark problems.

The rest of this paper is structured as follows. The next section gives a very brief overview of the most relevant background work.
Section~\ref{sec:multi} describes two variations of the compact Genetic Algorithm (cGA):
the Multi-Sample version and the Sliding Window version.
Both improve significantly on the standard cGA, with the sliding window version (a novel algorithm) providing the best results in our experiments.
Section~\ref{sec:problems} describes the test problems used and Section~\ref{sec:results} presents the results.  Section~\ref{sec:conc} concludes and also discusses ongoing and future work.

\section{Background}
\label{sec:background}

The cGA~\cite{harik1999compact} is an Estimation of Distribution Algorithm (EDA) for binary optimisation problems. In contrast to the standard Genetic Algorithm (GA)~\cite{mitchell1998introduction}, no cross-over or mutation is involved. Instead, the cGA uses a virtual population represented as a probability distribution over the set of binary strings. At each optimisation iteration, exactly two individuals are sampled from the distribution and evaluated for fitness to pick a winner and loser.  The probability distribution is then adjusted to increase the probability of generating the winning vector. The iterations continue until either the evaluation budget has been exhausted or the distribution has converged (such that the probability of producing a `1' in each position is either zero or one). 

The cGA performs well in noisy conditions, as clearly shown in Friedrich et al \cite{ExtremeNoiseCGA}. The main contribution of this paper is to develop two new versions of the cGA that make even more efficient use of the available evaluation budget. Both algorithms are efficient and easy to implement.

The standard cGA is a type of Univariate Estimation of Distribution Algorithm, since each dimension is considered independently of all others.
Harik emphasises that the algorithm can be sensitive to the probability model used to model the virtual population and that ``the choice of a good distribution is equivalent to linkage learning''~\cite{harik1999linkage}. Regarding this, the extended cGA (ECGA) using a set of probability models known as Marginal Product Models (MPMs) is proposed~\cite{harik1999linkage}. An MPM can represent a probability distribution over one bit or a tuple of bits, taking into account the dependency among the bits, and hence modelling higher-order effects.  A similar idea has been used in a so-called N-Tuple system by Kunanusont et al.~\cite{Kunanusont2017bandit}.

The multiple sample variations of the cGA introduced in this paper should also work with higher-order probability models, but this has not yet been implemented or tested.

\subsection{Compact Genetic Algorithm (cGA)}
In this section we describe the standard cGA in Alg. \ref{algo:cga}, as introduced in \cite{harik1999compact}.  
It models a population using a $d$-dimensional vector of probabilities, with an element for each bit of the solution.  The algorithm has one parameter $k$ which refers to the virtual population size and determines the learning rate, $\frac1k$.

Each element of the vector is initialised to $0.5$ and represents the probability that each corresponding value in the solution string should be a $1$.  At each iteration two random binary vectors are produced by sampling from the probability vector, and the fitness of each one is evaluated ($\Call{reproduceAndEvaluate}{\cdot}$).
The fitness values are compared to determine a winner and loser ($\Call{compete}{\cdot}$) and the probability distribution is updated: no update occurs if the two vectors have the same value. 
The algorithm then iterates over each dimension, comparing each candidate bit-by-bit. Updates to the probability vector only occur when the corresponding bits in the winner and loser differ.
If the winning bit is $1$, then the probability of producing a $1$ in that position is increased by $\frac1k$, otherwise (i.e.\ if the winning bit is a $0$) the probability is decreased by $\frac1k$, as defined in $\Call{update}{\cdot}$.

The algorithm terminates when the probability vector has converged as defined in function $\Call{isConverged}$. We also stop the algorithm when the total evaluation budget has been consumed. The solution found by the algorithm is the argmax of the probability vector, i.e. a $0$ whenever the corresponding probability is less than $0.5$ and a $1$ otherwise (cf. $\Call{recommend}$).

Note that this is a rank-based algorithm, in that the magnitude of the difference in fitness between winner and loser makes no difference.

The single parameter, virtual populations size $k$, has an important effect on the algorithm's behaviour. Setting $k$ too low (and hence the learning rate too high) causes premature convergence of the probability vector, and results in very poor final solutions. Setting $k$ too high causes slower than necessary convergence but does not harm solution quality so much. Friedrich et al \cite{ExtremeNoiseCGA} show how to set $k$ optimally when the noise model is Gaussian-distributed and the noise variance is known. For this paper we compare a range of $k$ values to the one used in \cite{ExtremeNoiseCGA}.

\begin{algorithm}[htbp]
\caption{\label{algo:cga}Standard compact Genetic Algorithm (cGA) for a $d$-bit binary problem.}
\begin{algorithmic}[1]
\Require{problem dimension $d$}
\Require{fitness function $fitness(\cdot): \{0,1\}^d \mapsto \R$}
\Require{virtual population size $k$}
    \Statex
    \Function{cGA}{$d,k$}
      \State{Initialise $p \gets \{\frac12\}^d$}
      \While{!\Call{isConverged}{$p$} }
        \State{$x_1,y_1 \gets $~\Call{reproduceAndEvaluate}{$p$}} 
        \State{$x_2,y_2 \gets $~\Call{reproduceAndEvaluate}{$p$}}
        \State{$x_{win}, x_{loss} \gets $~\Call{compete}{$x_1,y_1,x_2,y_2$}}
        \State{$p \gets $~\Call{update}{$p,x_{win}, x_{loss}$}}
      \EndWhile
      \State{$x^* \gets$~\Call{recommend}{$p$}}
      \State{\Return{$x^*$}}
    \EndFunction
    \Statex
    \Function{reproduceAndEvaluate}{$p$}
    	\State{$x \gets \{0\}^{|p|}$}
    	\For{$i \in \{1,\dots,d\}$}
        	\If{$rand < p(i)$} 
            	\State{$x(i) = 1$}
            \EndIf
        \EndFor
        \State{$y \gets fitness(x)$}
    	\State{\Return{$x,y$}}
    \EndFunction
        \Statex
    \Function{compete}{$x,y,x',y'$}
        \If{$y > y'$} 
            \State{$x_{win} \gets x$}
            \State{$x_{loss} \gets x'$}
        \Else
            \State{$x_{win} \gets x'$}
            \State{$x_{loss} \gets x$}
        \EndIf
    	\Return{$x_{win}, x_{loss}$}
    \EndFunction
    \Statex
    \Function{update}{$p, x_{win}, x_{loss}$}
    	\For{$i \in \{1,\dots,|d|\}$}
          \If{$x_{win}(i) \neq x_{loss}(i)$} 
              \If{$x_{win}(i)== 1$} 
                  \State{$p(i)=p(i) + \frac1k$} 
              \Else
                  \State{$p(i)=p(i) - \frac1k$} 
              \EndIf
          \EndIf
        \EndFor
        \Return{$p$}
    \EndFunction
    \Statex
    \Function{recommend}{$p$}
    	\State{$x \gets \{0\}^{|p|}$}
    	\For{$i \in \{1,\dots,d\}$}
        	\If{$p(i)> \frac12$} 
            	\State{$x(i) = 1$}
            \EndIf
        \EndFor
    	\Return{$x$}
    \EndFunction
        \Statex
    \Function{isConverged}{$p$}
    	\For{$i \in \{1,\dots,d\}$}
        	\If{$0<p(i)<1$} 
            	\State \Return{$false$}
            \EndIf
        \EndFor
    	\State \Return{$true$}
    \EndFunction
  \end{algorithmic}
\end{algorithm}

\section{Multiple-Sample cGA Variants}
\label{sec:multi}
This section describes the novel contribution of the paper. 

The motivation for the new algorithm is to make better use of the fitness evaluations in order to find optimal or close to optimal solutions more quickly. Observe that in the standard cGA, at each iteration we draw two samples from the distribution and make one comparison and one update of the probability vector.  
This gives us an update per sample ratio of $1/2$. The question arises as to whether we can make more efficient use of the samples we evaluate (keeping in mind that for most practical applications, the main cost of an evolutionary algorithm is in performing fitness evaluations).

This observation raises the question as to whether we may make better use of the fitness evaluations if we make more comparisons and updates for each one.  This leads us on to the following two algorithms. The first is a natural
extension of the cGA to increase the number of individuals sampled at each iteration.  Note that this algorithm was described in \cite{harik1997compact}. The second version
only samples and evaluates a single candidate solution at each iteration, but then makes comparisons and updates with a number of previously evaluated vectors stored in a sliding history window.

\subsection{Multiple-Sample per Iteration cGA}

\begin{algorithm}[htbp]
\caption{\label{algo:mscga}Multiple-Sample compact Genetic Algorithm (MScGA) for a $d$-bit binary problem. Note that the standard cGA is the special case of MScGA with sample number $n=2$.}
\begin{algorithmic}[1]
\Require{problem dimension $d$}
\Require{fitness function $fitness(\cdot): \{0,1\}^d \mapsto \R$}
\Require{virtual population size $k$}
\Require{sample number $n$}
    \Statex
    \Function{MScGA}{$d,k,n$}
      \State{Initialise $p \gets \{\frac12\}^d$}
      \While{! \Call{isConverged}{$p$}}
      	\For{$i \in \{1,\dots,n\}$}
        	\State{$x_i, y_i \gets $~\Call{reproduceAndEvaluate}{$p$}} 
        \EndFor
        \State{$j_1,j_2,\dots,j_n\gets $~\Call{rankDescend}{$y_1,y_2,\dots,y_n$}}
        \For{$a \in \{1,\dots,n-1\}$}
        	\For{$b \in \{a+1,\dots,n\}$}
              \State{$x_{win} \gets x_{j_a}$} 
              \State{$x_{loss} \gets x_{j_b}$}
              \State{$p \gets $~\Call{update}{$p,x_{win}, x_{loss}$}}
        	\EndFor
      	\EndFor
      \EndWhile
      \State{$x^* \gets$~\Call{recommend}{$p$}}
      \State{\Return{$x^*$}}
    \EndFunction
        \Statex
    \Function{rankDescend}{$y_1,y_2,\dots,y_n$}
        \State{Define $j_1, j_2,\dots, j_n$ such that $y_{j_1} \geq y_{j_2} \geq \dots \geq y_{j_n}$}
    	\State{\Return{$j_1, j_2, \dots,j_n$}}
    \EndFunction
  \end{algorithmic}
\end{algorithm}

In the multi-sample version we now make $n$ samples per iteration and present it in Alg.~\ref{algo:mscga}.
Apart from this detail, the algorithm is very similar to the 
standard cGA.
Since we are now making $n$ samples and $n$ evaluations, we now have $n(n-1)/2$ comparisons and updates to make.  For instance, for $n=10$ the ratio of updates per sample is now $4.5$, nine times higher than the standard $n=2$ case.

Note that an algorithm similar to this was described in \cite{harik1997compact} though the way the algorithm was listed did not separate the fitness evaluation from the comparison (which is necessary in order to make
best use of the fitness evaluation budget), though this detail may have been considered to be a low-level implementation detail by the authors. More importantly, the results presented in \cite{harik1997compact} for the multi-sample case were not particular good, perhaps due
to a poorly chosen $k$ value. When making more updates per fitness evaluation, $k$ needs to be set higher to avoid premature convergence.

This could be the reason why recent work on the cGA \cite{ExtremeNoiseCGA} has not mentioned the Multiple-Sample variant.  We will show that when $k$ is chosen well, the Multiple-Sample cGA (MScGA) greatly outperforms the standard cGA.

\subsection{Sliding Window cGA}

\begin{algorithm}[htbp]
\caption{\label{algo:swcga}Sliding Window compact Genetic Algorithm (SWcGA) for a $d$-bit binary problem.}
\begin{algorithmic}[1]
\Require{problem dimension $d$}
\Require{fitness function $fitness(\cdot): \{0,1\}^d \mapsto \R$}
\Require{virtual population size $k$}
\Require{sliding window width $w$}
    \Statex
    \Function{cGA}{$d,k,w$}
      \State{Initialise $p \gets \{\frac12\}^d$}
      \State{Initialise empty queue $QX$ to save individuals}
      \State{Initialise empty queue $QY$ to save fitness values}
      \While{! \Call{isConverged}{$p$}}
        \State{$x,y  \gets $~\Call{reproduceAndEvaluate}{$p$}} 
        \For{$h \in \{1,\dots,length(QX)\}$}
          \State{$x_{win}, x_{loss} \gets $~\Call{compete}{$x,y,QX_h,QY_h$}}
          \State{$p \gets $~\Call{update}{$p,x_{win}, x_{loss}$}}
        \EndFor
        \State{$QueueX \gets$~\Call{FIFO}{$x,QX,w$}}
        \State{$QueueY \gets$~\Call{FIFO}{$y,QY,w$}}
        \EndWhile
      \State{$x^* \gets$~\Call{recommend}{$p$}}
      \State{\Return{$x^*$}}
    \EndFunction
    \Function{compete}{$x,y,x',y'$}
        \If{$y > y'$} 
            \State{$x_{win} \gets x$}
            \State{$x_{loss} \gets x'$}
        \Else
            \State{$x_{win} \gets x'$}
            \State{$x_{loss} \gets x$}
        \EndIf
    	\Return{$x_{win}, x_{loss}$}
    \EndFunction
    \Statex
    \Function{FIFO}{$e,Q,w$}
    	\If{$length(Q) == w$} 
	\State{Dequeue the first element of $Queue$}
        \EndIf
         \State{Enqueue $e$ to the tail of $Queue$}
    	\State{\Return{$Queue$}}
    \EndFunction
  \end{algorithmic}
\end{algorithm}
While the MScGA aims to provide more efficient use of the available fitness evaluations, it suffers from the fact that
the probability vector is only updated after all the samples for an iteration have been drawn.

However, it may be beneficial to update the probability vector more frequently, ideally after every new sample has been drawn and evaluated.
This is exactly what the Sliding Window cGA (SWcGA) achieves.  In addition to the parameter $k$, this algorithm adds the parameter $w$ for the size of the window.

Again, the algorithm is similar to the standard cGA, except that now every time a sample is drawn from the probability vector, the fitness is evaluated and then the scored vector is compared with every other one in the window, and for each comparison the probability vector is updated.
Note that each sample only has its fitness evaluated once and stored together with the sample in the sliding window (which can be implemented as a circular buffer or a FIFO queue).  
See algorithm~\ref{algo:swcga} for the listing.  For each new candidate sampled, assuming steady state when the buffer is already full, we make comparisons and updates with the $w$ previously evaluated samples.
Hence, the ratio of comparisons and updates to fitness evaluations is $w$.
After the comparisons and updates have been made, the new scored sample is added to the sliding window buffer, replacing the oldest one if the buffer is already full (i.e. already has $w$ scored samples in it).

\section{Test Problems}
\label{sec:problems}

We considered two binary optimisation problems based on bit strings: the OneMax problem corrupted by additive Gaussian noise, namely noisy OneMax, and the noisy PMax problem.  

\subsection{Noisy OneMax}
The OneMax problem aims at maximising the number of $1$ bits in a binary string.
Let $\G(\mu,\sigma^2)$ denote Gaussian noise with mean $\mu$ and variance $\sigma^2$, and $\x$ is an $d$-bit binary string.
The $d$-bit OneMax problem with additive Gaussian noise is formalised as $f(\x)=\sum_{i=1}^{d} \x_i + \G(0,1)$.
Friedrich et al~\cite{ExtremeNoiseCGA} have proven that with high probability, the standard cGA with $k =\omega(\sigma 2\sqrt{d}\log{d})$ converges to the optimal distribution $p^*$ after $O(k\sigma^2\sqrt{d}\log{kd})$ iterations when optimising a noisy OneMax with variance $\sigma^2>0$, where $k=7\sigma^2\sqrt{d}(\ln{d})^2$ is used.
Thus, in the case considered in this paper ($\sigma^2=1$, $d=100$), the $k$ should be $7*\sqrt{d}(\ln{d})^2 \approx 15d $ to guarantee the convergence. This setting is compared to as baselines in our experiments.

\subsection{Noisy PMax}
The Noisy PMax problem is proposed by Lucas et al.~\cite{lucas2017evaluation} to represent an artificial game outcome optimisation problem.
In this artificial model, $\x$ is treated as an $d$-bit binary number, and the true winning rate of $\x$ is defined as $\P_{win}(\x) = \frac{Value(\x)}{2^d-1}$,
where $Value(\x)$ denotes the numeric value of $\x$ located between $0$ and $(2^d-1)$.
Thus, the outcome of a game is either win ($1$) with probability $\P_{win}(\x)$ or loss, otherwise.  

This problem formulation is also relevant to learning playout control parameters for Monte Carlo Tree Search (MCTS)~\cite{Cazenave2016Playout}, where the parameters control the biases for a stochastic playout policy.  The efficient cGA variants described in this paper should be able to improve on the simple evolutionary algorithms used in \cite{Lucas2014FastEvoAdapt,Perez2014KnowledgeBased} but this has not yet been tried.  The relevance is due to three factors: the extreme noise when evaluating stochastic playout policies, and the requirement for rapid adaptation (a feature of the MScGA algorithms), and the expectation that different parameters have very different levels of importance in controlling the playouts.

\section{Experimental results and discussion}
\label{sec:results}

\subsection{Experimental setting}
We consider two baseline algorithms, the standard cGA and the Random Mutation Hill Climber (RMHC) on each of the tested problems.
For each experimental run each algorithm was given a fixed maximum budget of $1,000$ fitness evaluations.  
Thus, the cGA and its variants stopped when the stopping condition defined in Alg. \ref{algo:cga} was met, or the (noisy) fitness function had been evaluated $1,000$ times.
Note that we did not use first hitting time as a measure, 
since this has been shown to give misleading results for noisy optimisation problems \cite{lucas2017evaluation}.
Since each algorithm under test is able to return its
best guess (by $\Call{recommend}{$p$}$ used by cGA and its variants) or best solution found so far (RMHC) at any iteration, we plotted the true (noise-free) fitness of the current solutions of each algorithm at each iteration.  Hence, in addition to the final fitness found we may also observe how fitness evolves over time.

First, we optimise separately the problems using cGA with different virtual population size $k$ and using RMHC with different resampling number $r$, then choose the $k$ and $r$ with the best performance, respectively. More study on the optimal resampling using RMHC on the OneMax with additive Gaussian noise can by found in \cite{liu2016optimal}.

\subsection{Noisy OneMax}

\begin{figure*}[htbp]
\centering
\begin{subfigure}[t]{1\columnwidth}\centering
\includegraphics[width=.8\columnwidth]{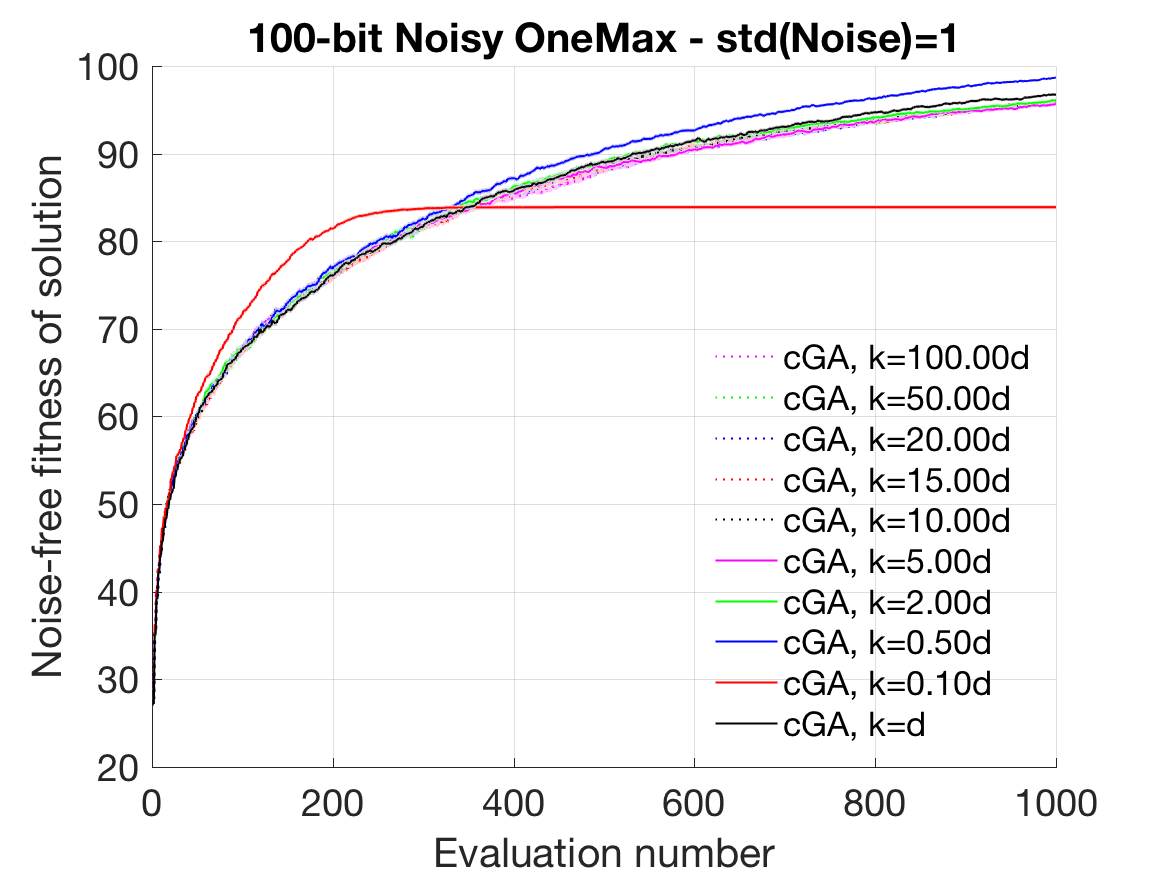}
\caption{\label{fig:baselineonemaxcga}Quality of solution recommended by the standard cGA with different learning rate $k$ using different budget. The performance of the standard cGA is less sensitive to the tested $k$ value on the tested problem.}
\end{subfigure}\hfill
\begin{subfigure}[t]{1\columnwidth}\centering
\includegraphics[width=.8\columnwidth]{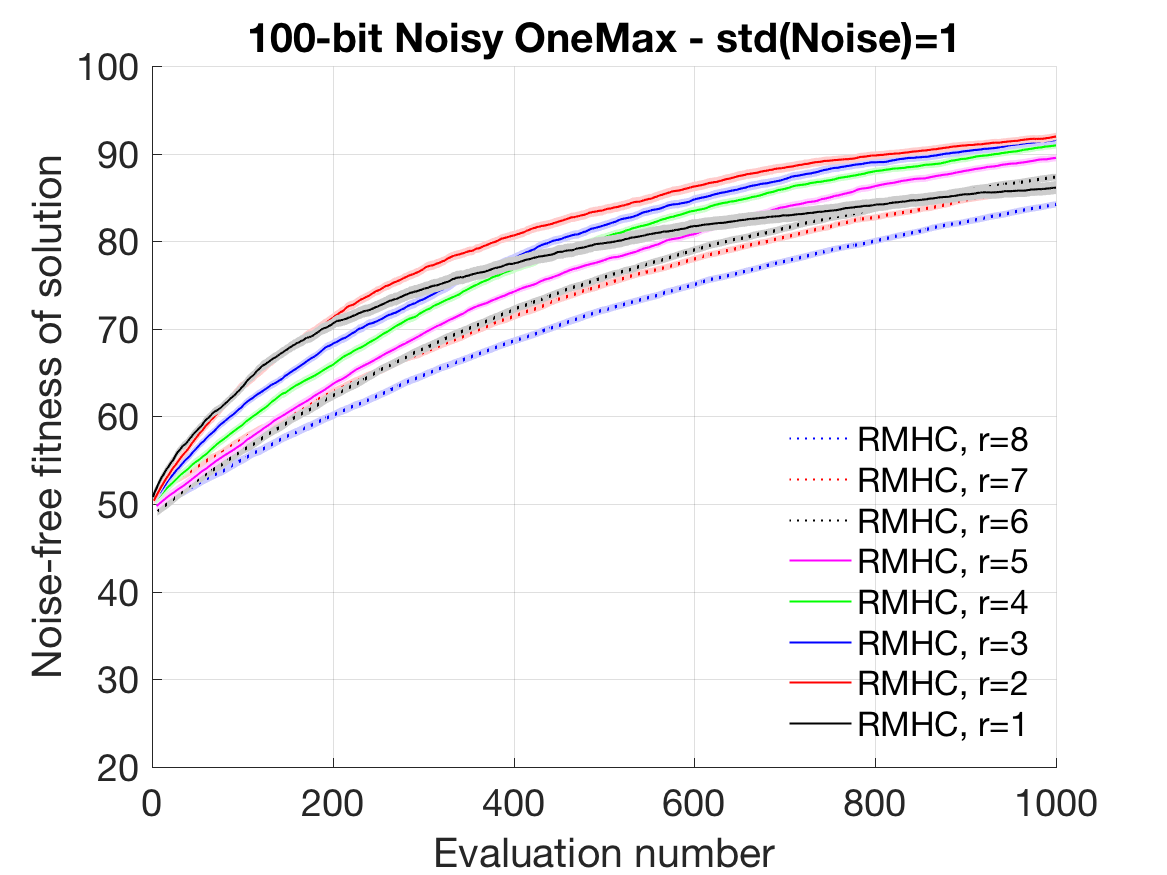}
\caption{\label{fig:baselineonemaxrmhc}Quality of solution recommended by RMHC with resampling number $r$ using different budget. The performance of RMHC without resampling (case $r=1$) is poor as predicted. The optimal resampling number for the tested problem is $3$ for bigger budget and $2$ for smaller budget.}
\end{subfigure}
\caption{\label{fig:baseline}Results of cGA and RMHC on the noisy OneMax problem. Each curve is an average of 100 trials, with $d = 100$. The standard error is also given as a faded area around the average.}
\end{figure*}

The performance of the standard cGA with different virtual population size $k$ and RMHC with different resampling number $r$ on the $100$-bit noisy OneMax problem is illustrated in Fig. \ref{fig:baseline}.
Fig. \ref{fig:baselineonemaxcga} shows that the cGA with virtual population size $k=d/10$ performs the best when the maximal fitness evaluation number is larger than $300$. It is notable that using small virtual population size, the cGA converges quickly to a good solution at the early stage of optimisation then never finds the optimum. As the RMHC with variant resampling numbers (see Fig. \ref{fig:baselineonemaxrmhc}) does not outperform the standard cGA with best $k=d/2$ (blue curve in Fig. \ref{fig:baselineonemaxcga}), our algorithms are directly compared to the standard cGA with $k=d/2$.

\begin{figure*}[htbp]
\centering
\includegraphics[width=.32\linewidth]{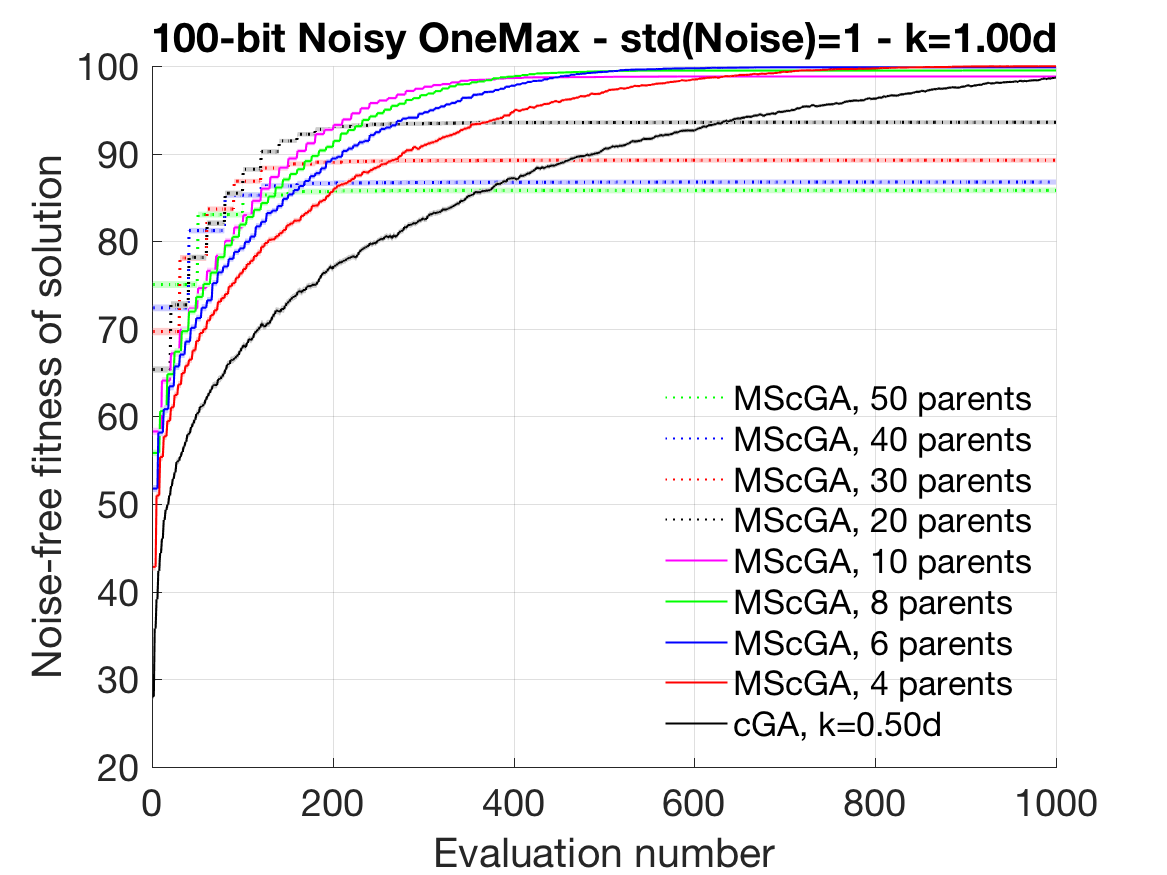}
\includegraphics[width=.32\linewidth]{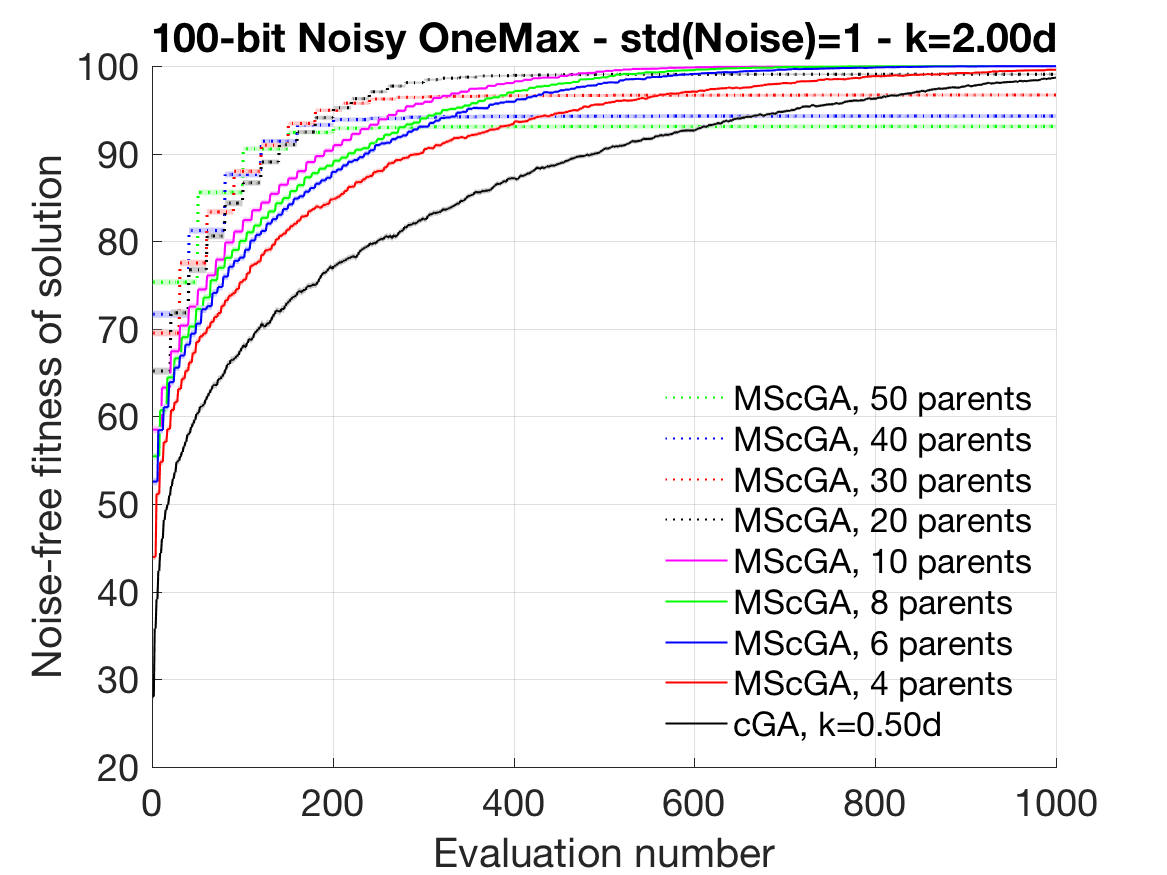}
\includegraphics[width=.32\linewidth]{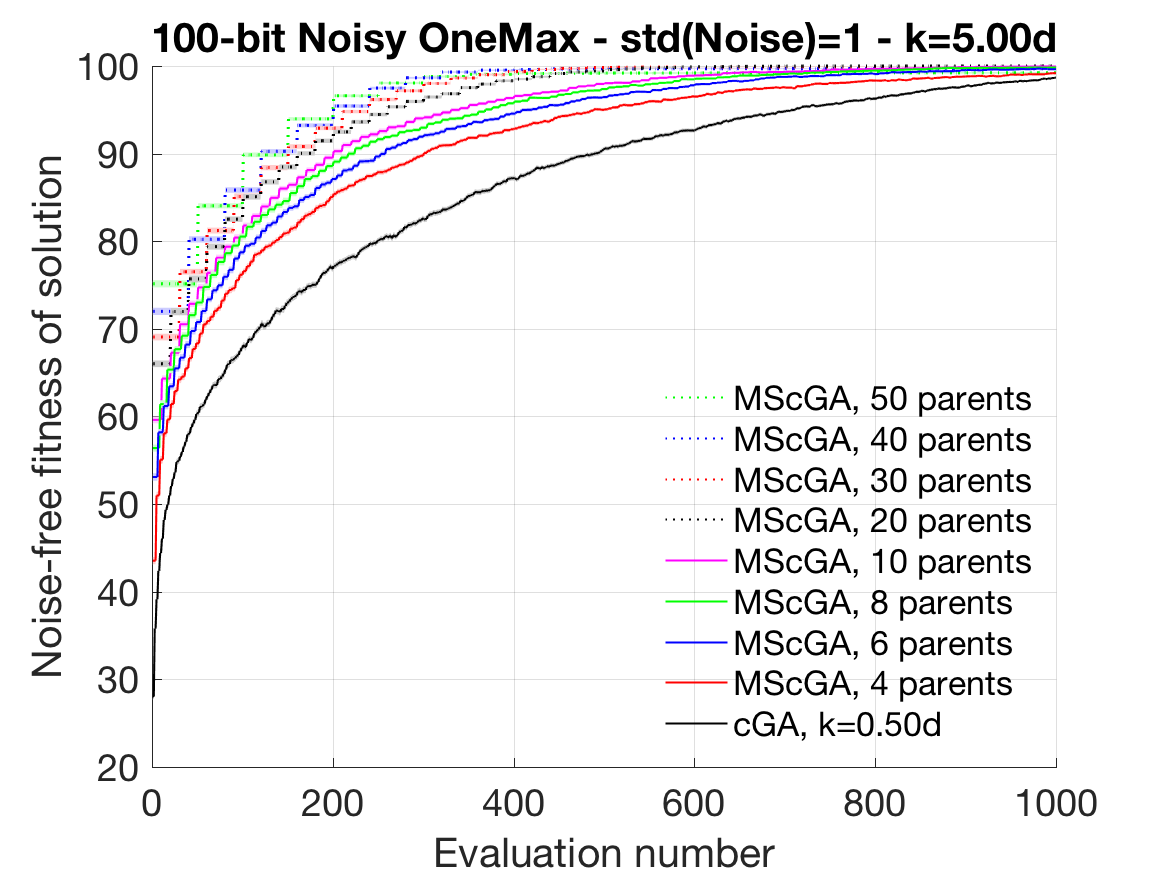}\\
\includegraphics[width=.32\linewidth]{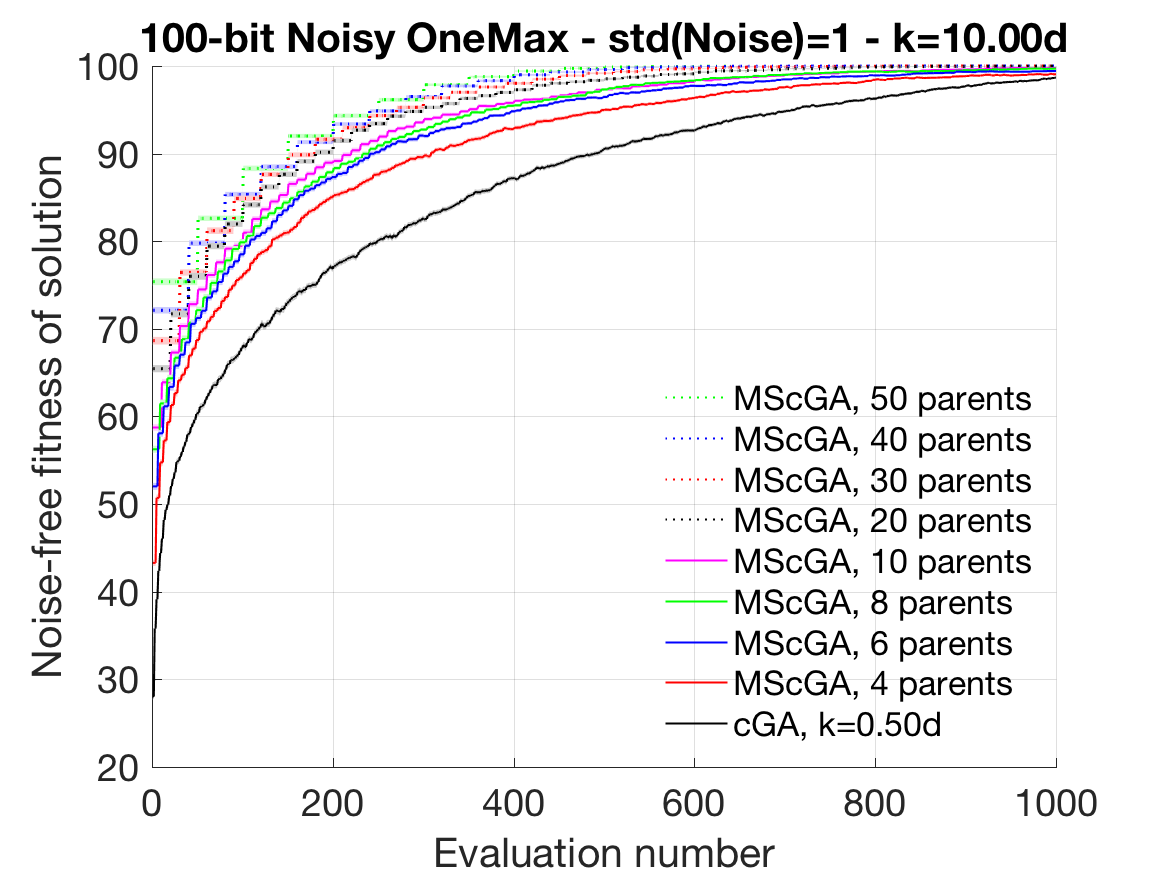}
\includegraphics[width=.32\linewidth]{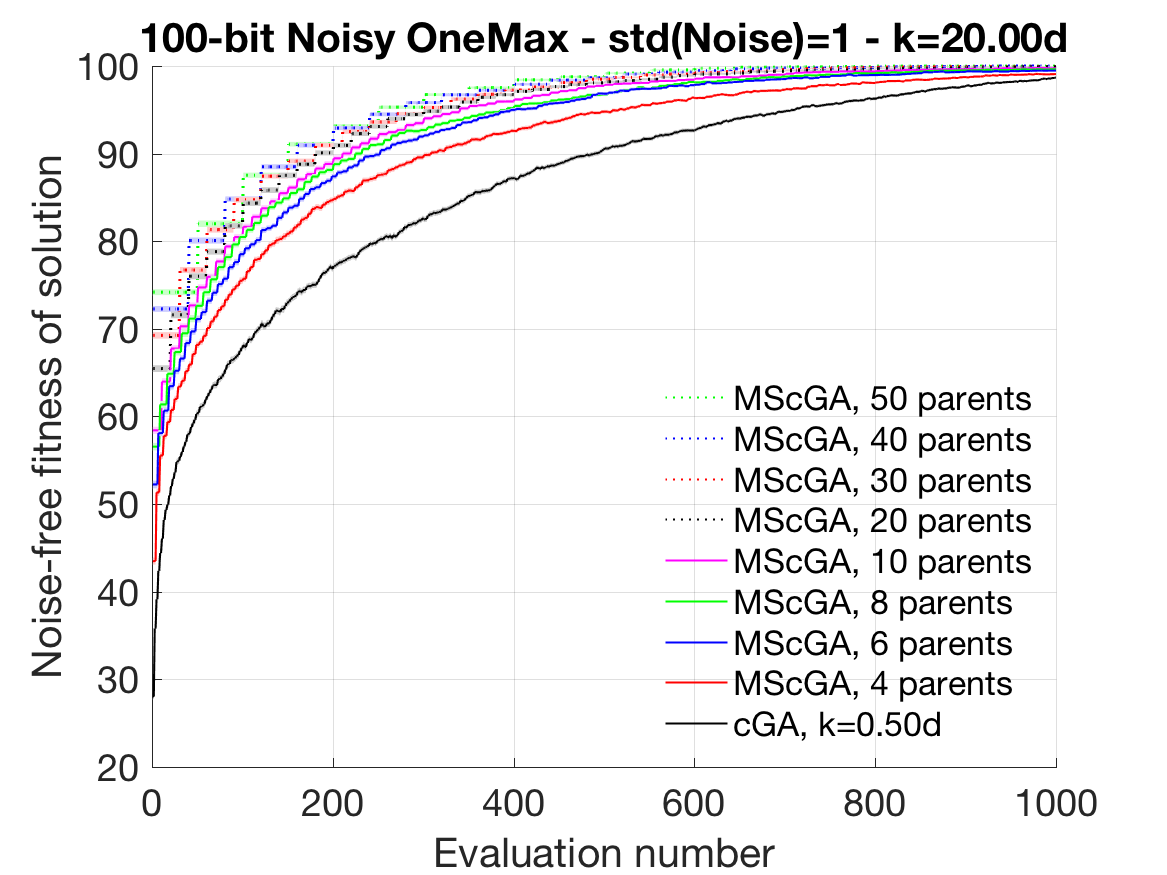}
\includegraphics[width=.32\linewidth]{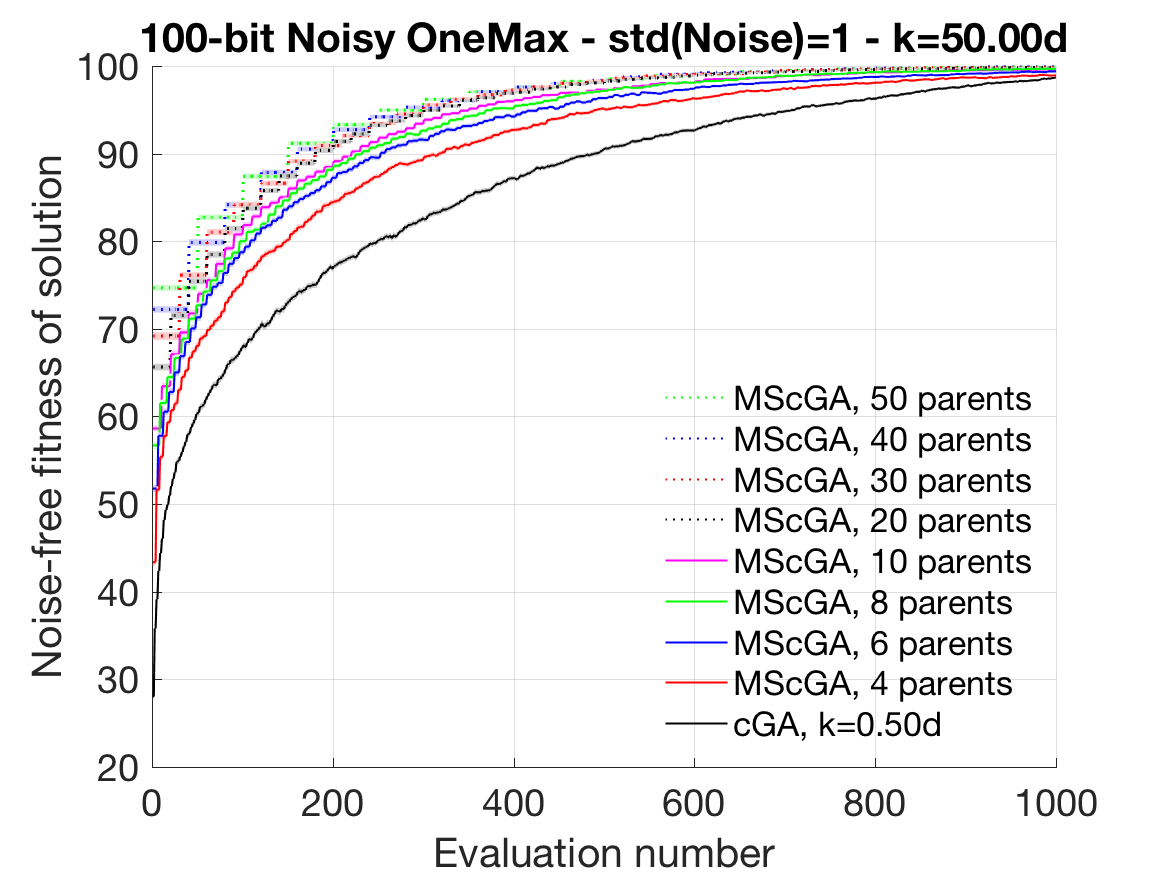}
\caption{\label{fig:mscgaonemax}Results of the noisy OneMax problem optimised by MScGA. Each curve is an average of 100 trials, with $d = 100$. The standard error is also given as a faded area around the average. The results using other values of $k$ are not shown as they are similar or worse than the ones shown.}
\end{figure*}

\begin{figure*}[htbp]
\centering
\includegraphics[width=.32\linewidth]{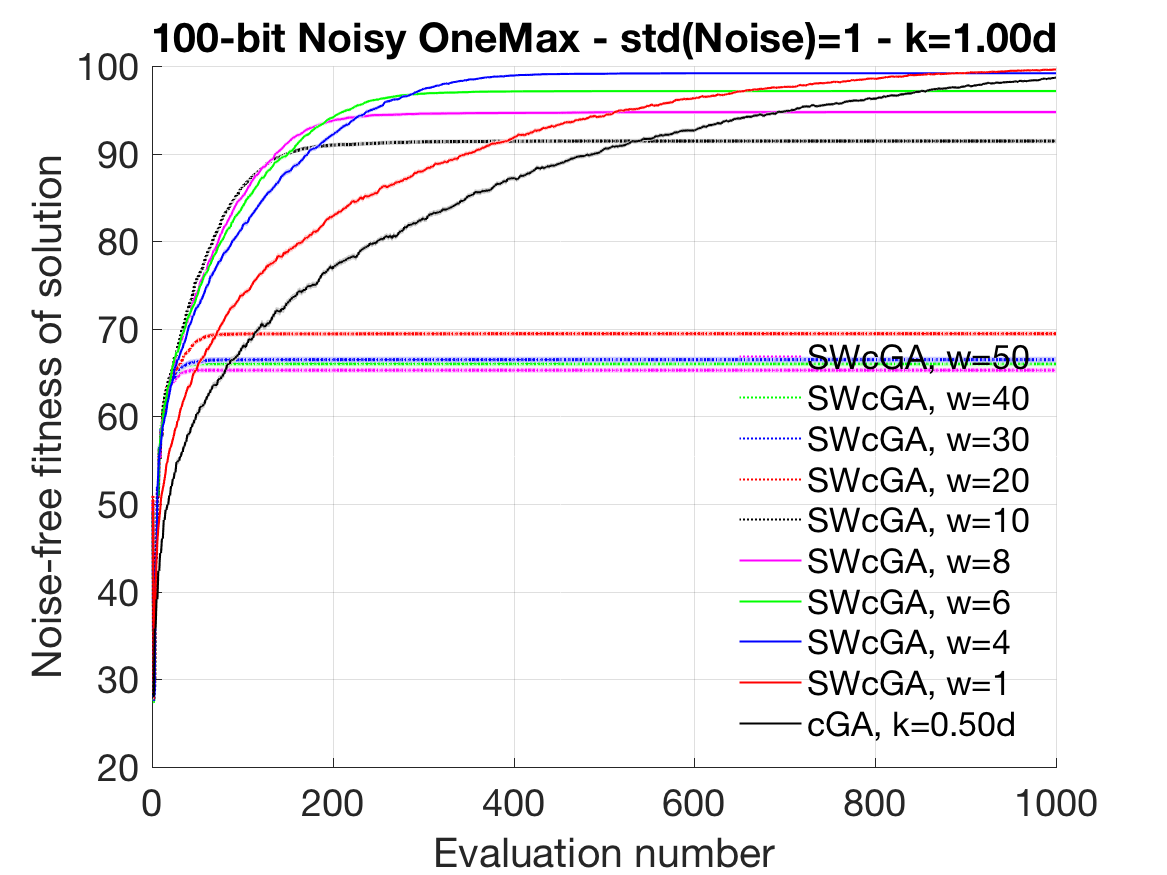}
\includegraphics[width=.32\linewidth]{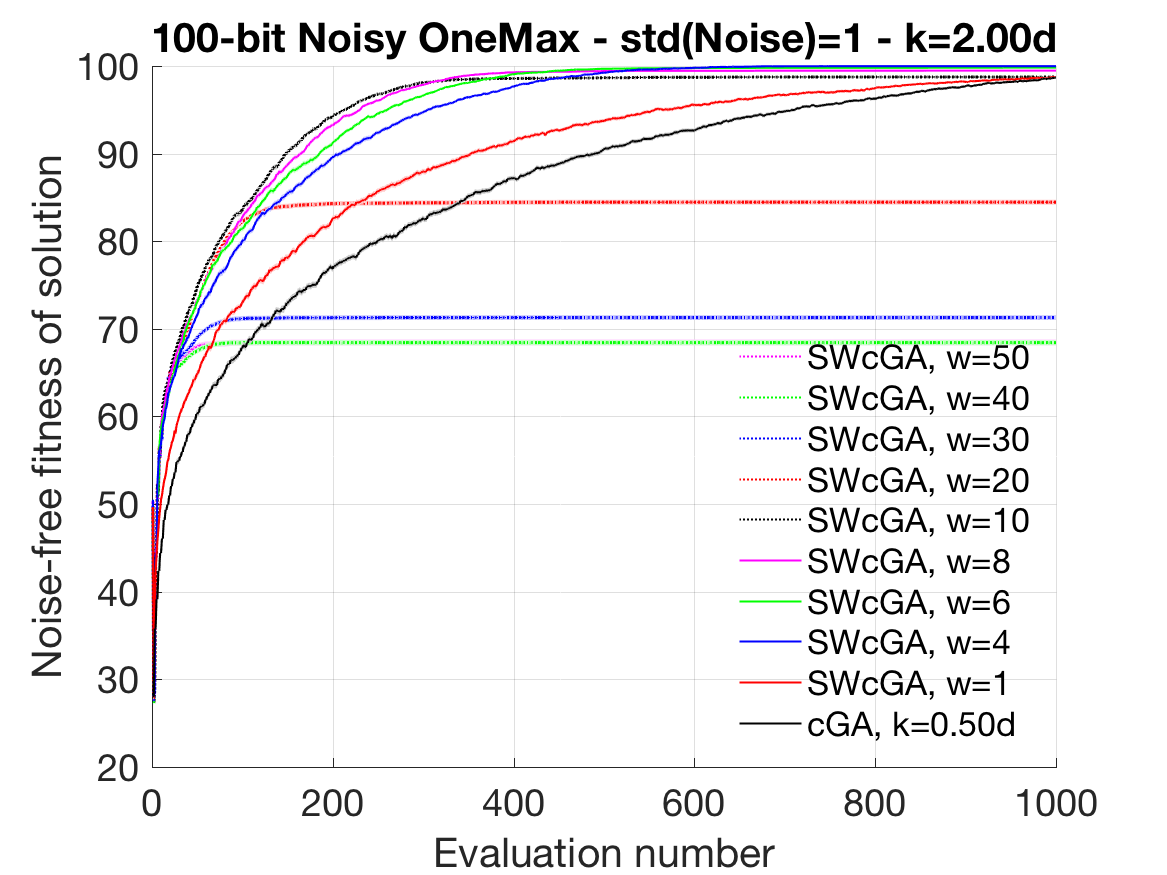}
\includegraphics[width=.32\linewidth]{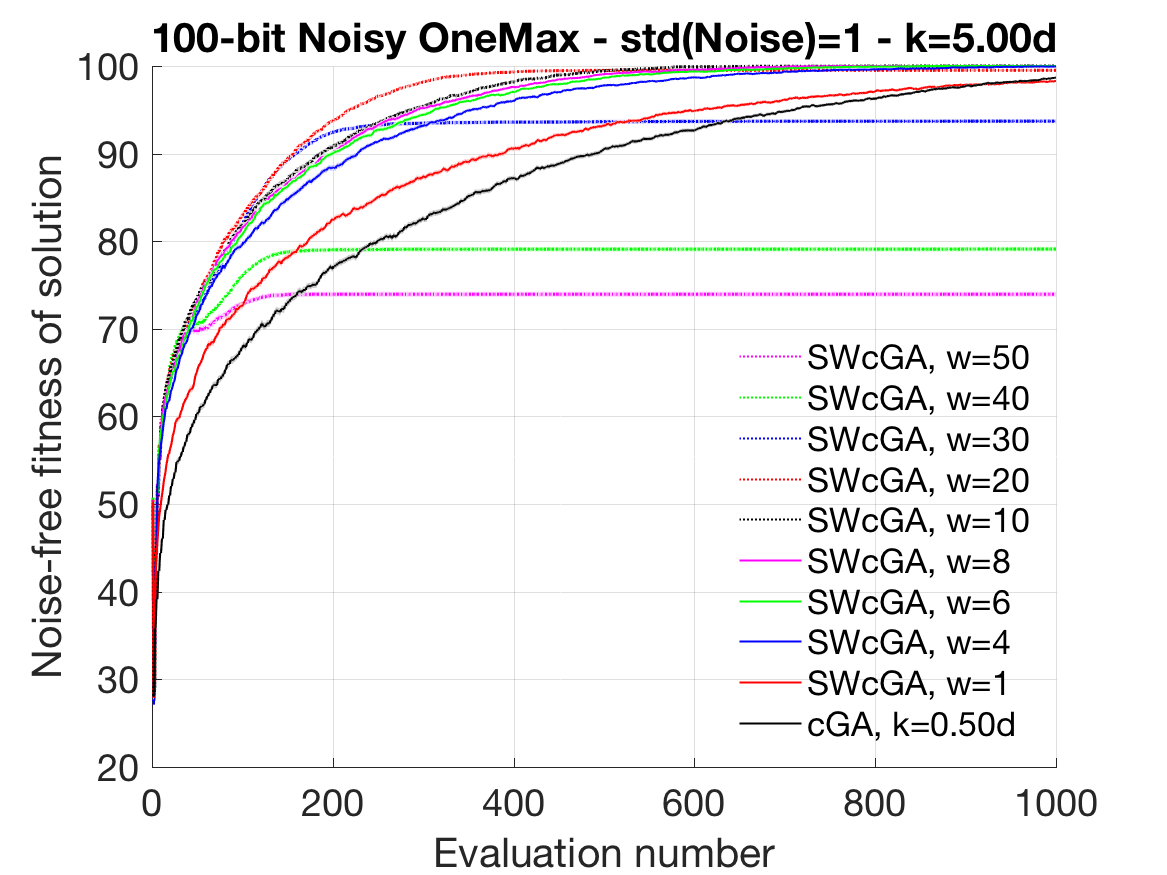}\\
\includegraphics[width=.32\linewidth]{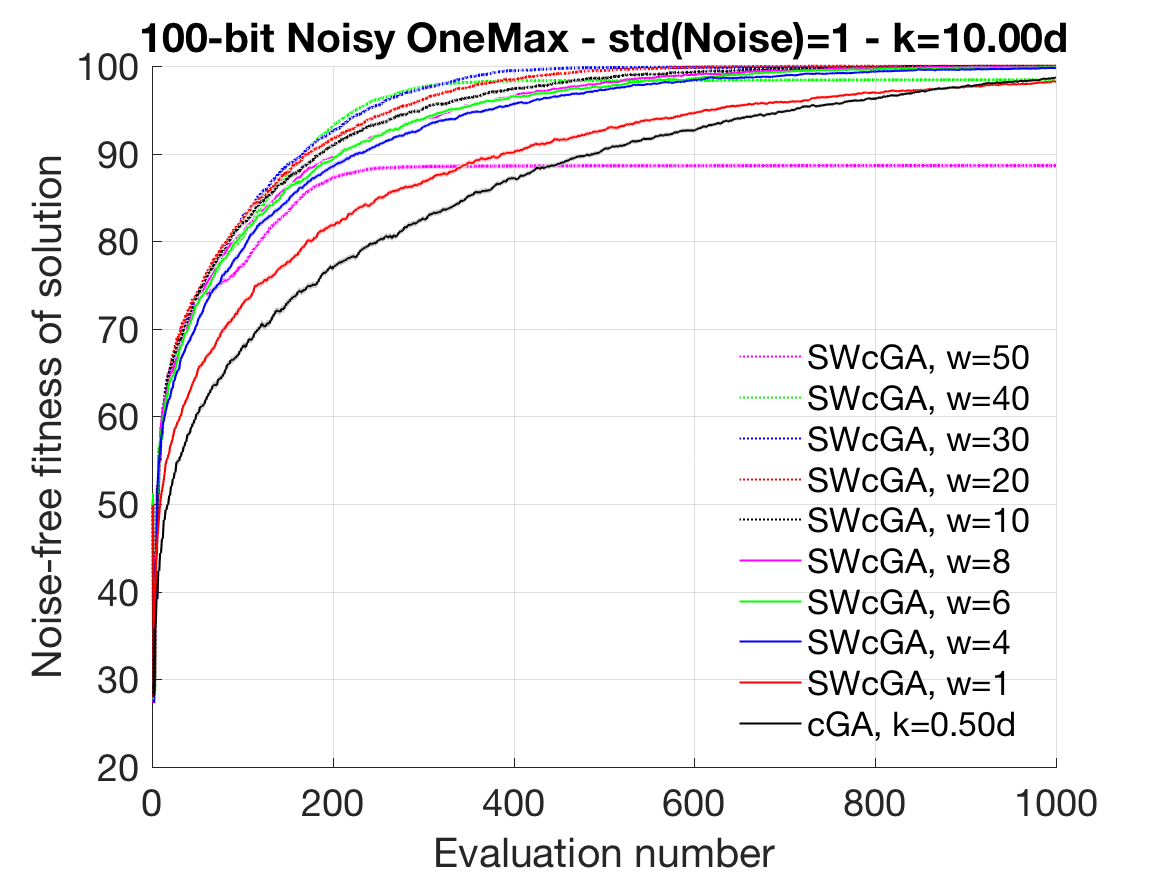}
\includegraphics[width=.32\linewidth]{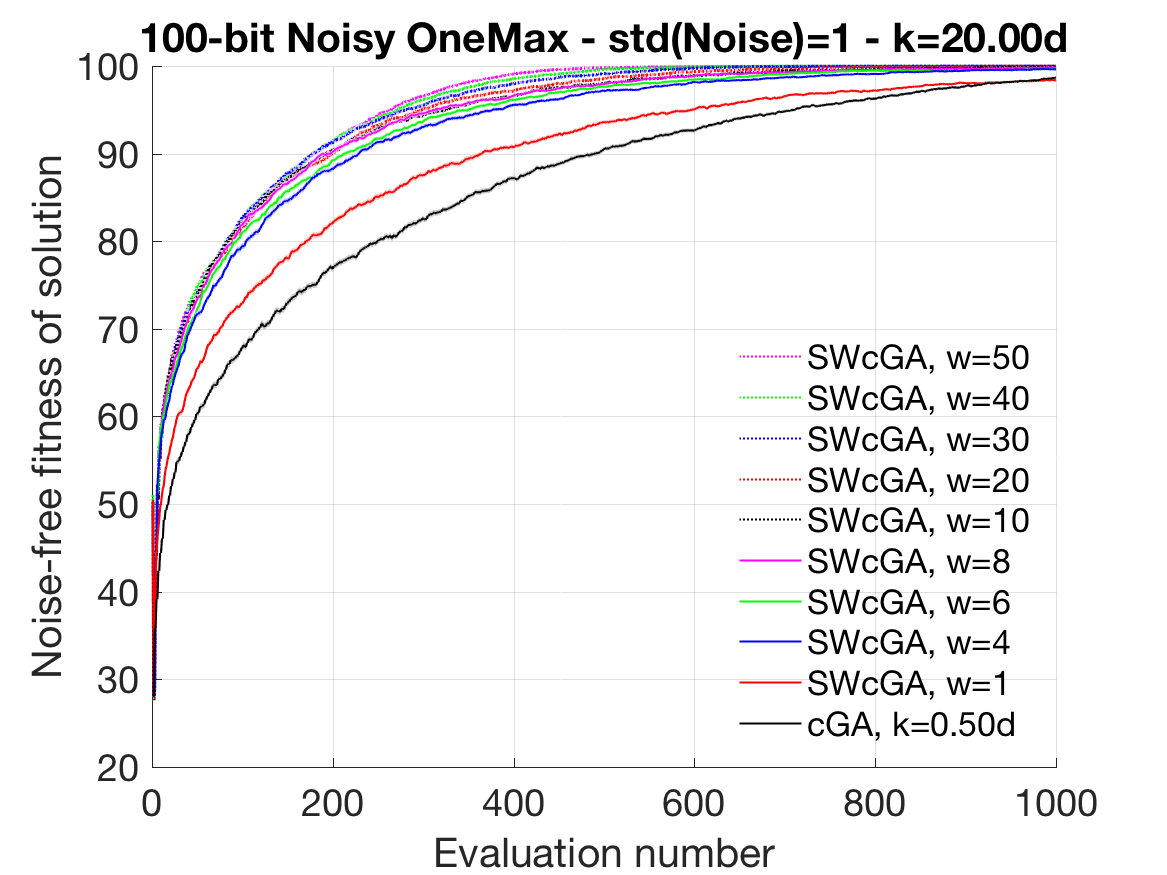}
\includegraphics[width=.32\linewidth]{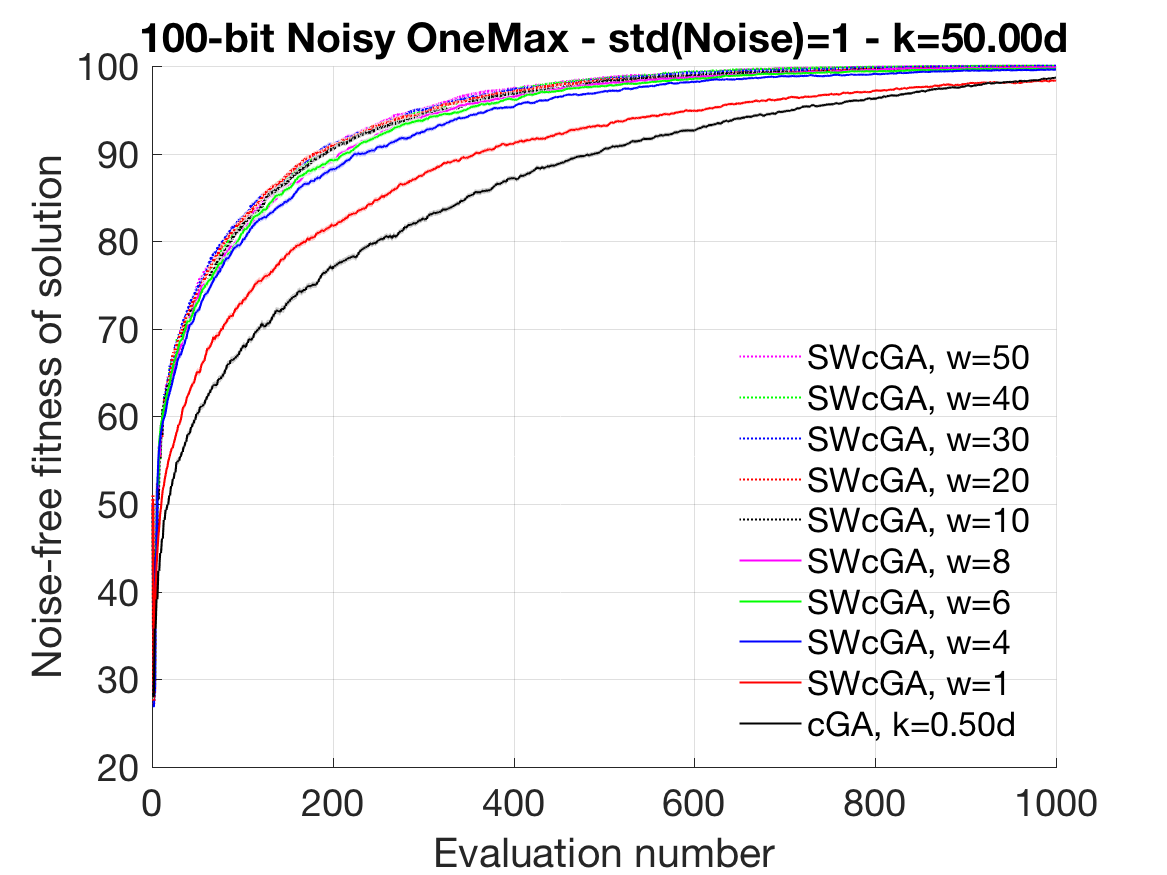}
\caption{\label{fig:swcgaonemax}Results of the noisy OneMax problem optimised by SWcGA. Each curve is an average of 100 trials. The standard error is also given as a faded area around the average. The results using other values of $k$ are not shown as they are similar or worse than the one shown.}
\end{figure*}

Fig. \ref{fig:mscgaonemax} compares the performance of MScGA instances using different virtual population size $k$ and sampling number $n$ and Fig. \ref{fig:swcgaonemax} compares the performance of SWcGA instances using different virtual population size $k$ and sliding window width $w$ on the identical noisy OneMax problem. 
When $k$ is close to $d$, the more samples there are, the worse the solution recommended by MScGA at each iteration is; the wider the sliding window is, the worse the solution recommended by SWcGA at each iteration is.
When $k$ is large, larger sample number leads to better performance of MScGA and MScGA significantly outperforms the best standard cGA, but the difference led by using different sample number is minor; wider window leads to better performance of SWcGA, and the overall performance is better than MScGA.
However, very big $k$ will weaken the performance of both MScGA and SWcGA. 
MScGA and SWcGA with optimal parameter setting have similar performance, but MScGA is less sensitive to its parameter, sample number $n$.


\begin{figure*}[htbp]
\centering
\begin{subfigure}[t]{1\columnwidth}\centering
\includegraphics[width=.8\columnwidth]{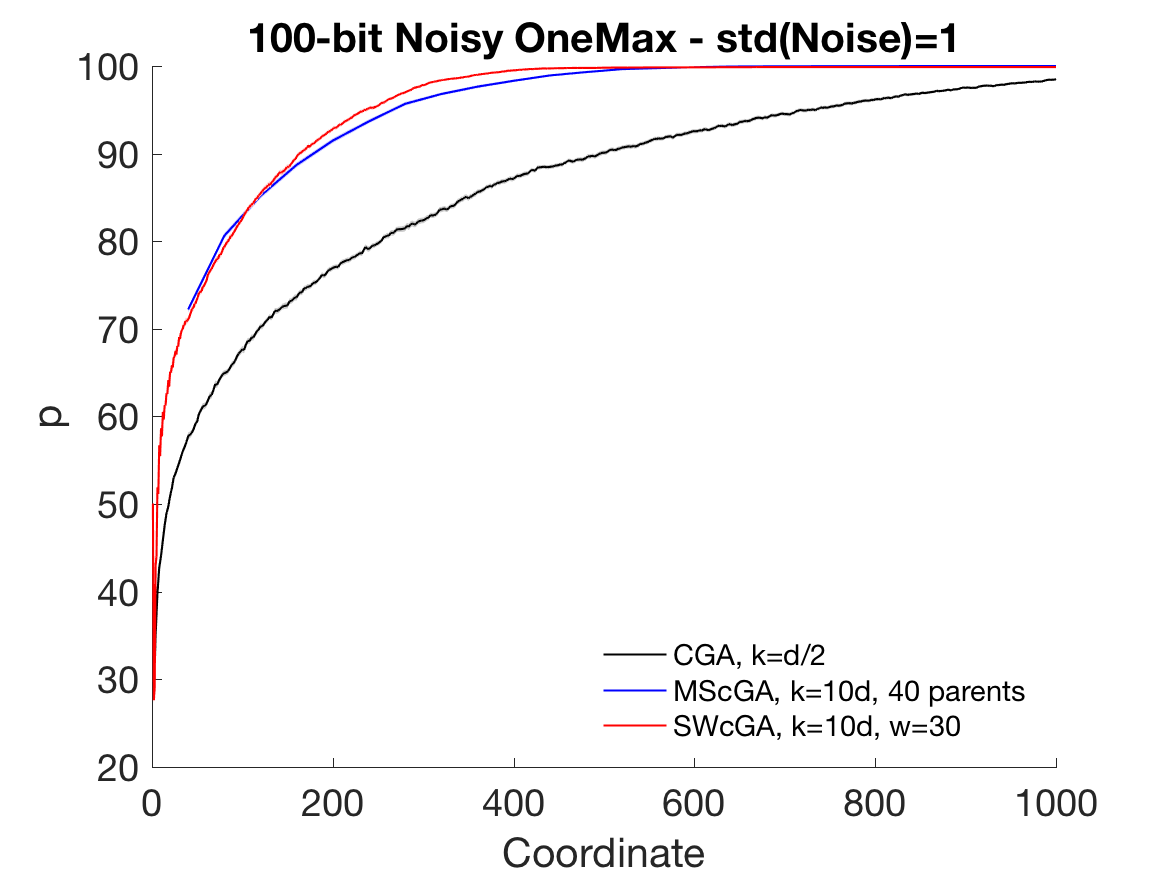}
\caption{Quality of recommendations.}
\end{subfigure}\hfill
\begin{subfigure}[t]{1\columnwidth}\centering
\includegraphics[width=.8\columnwidth]{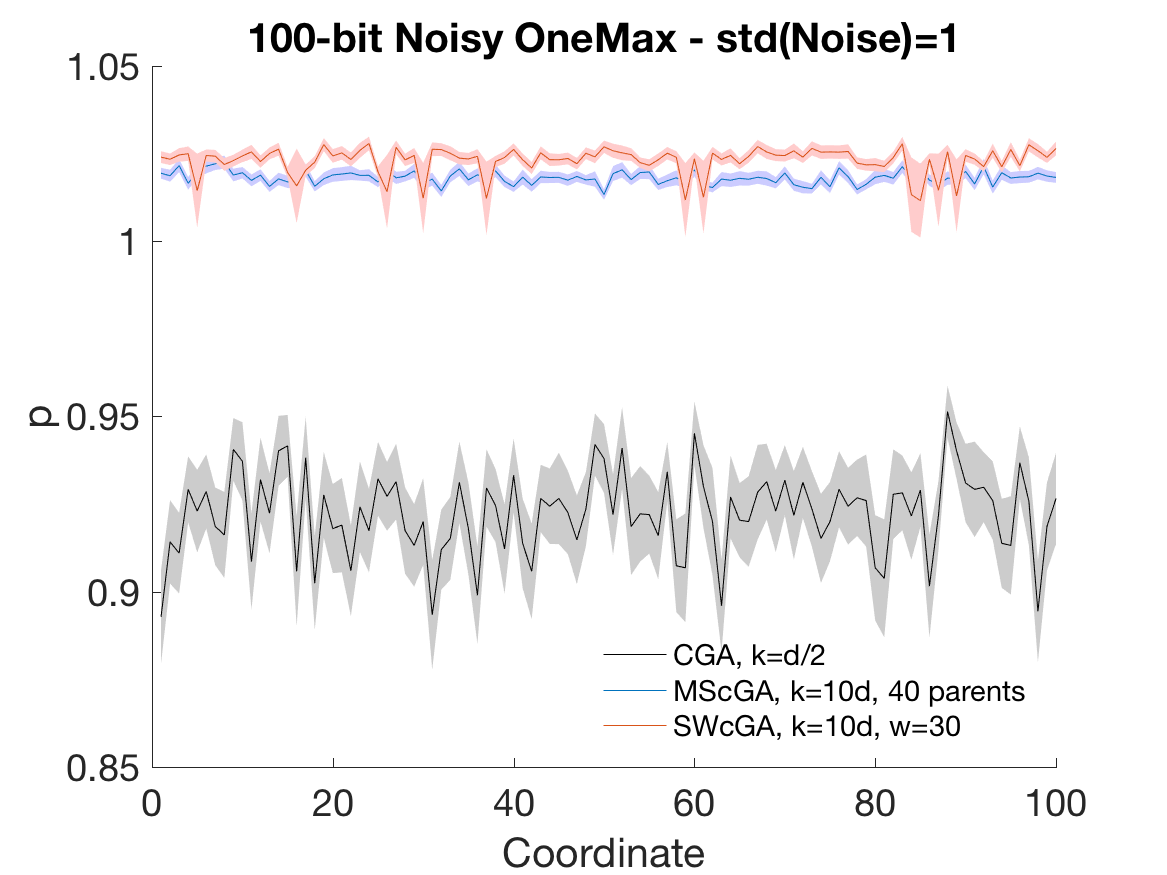}
\caption{Averaged final probability vector $p$.}
\end{subfigure}
\caption{\label{fig:best}Results of the noisy OneMax problem optimised by the cGA, MScGA and WScGA with best tested parameter settings. Each curve is an average of 100 trials. The standard error is also given as a faded area around the average.}
\end{figure*}

The best parameter settings of each of the algorithms are listed and compared in Fig. \ref{fig:best}, as well as the averaged final probability vector $p$ over 100 trials. Though MScGA converges faster than SWcGA, it did not stop with better solutions than SWcGA. The averaged noise-free fitnesses of the final recommendations are 98.68 ($\pm$ 0.12) by cGA, 100.00 ($\pm$ 0.00) by MScGA and 100.00 ($\pm$ 0.00) by SWcGA.

\subsection{PMax}
\begin{figure*}[htbp]
\centering
\begin{subfigure}[t]{1\columnwidth}\centering
\includegraphics[width=.8\columnwidth]{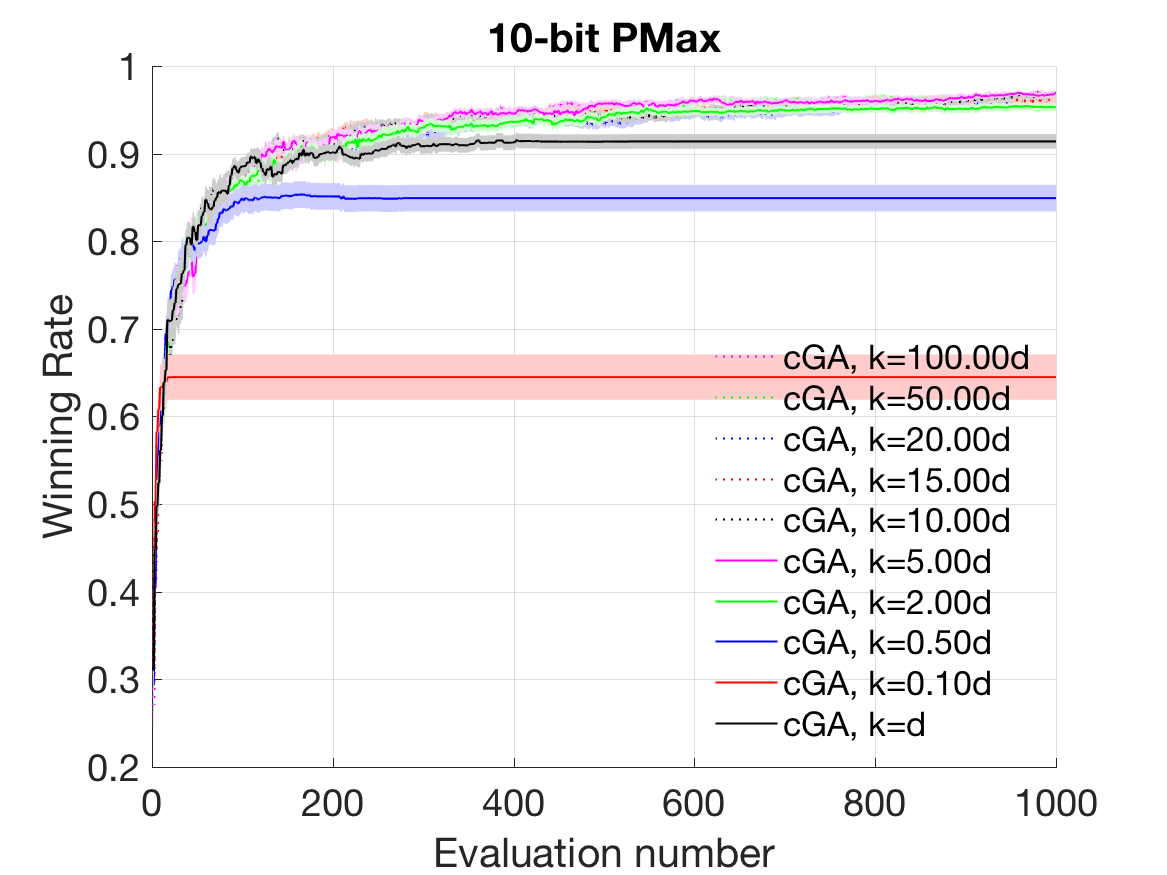}
\caption{\label{fig:baselinepmaxcga}Quality of solution recommended by the standard cGA with different virtual population size $k$ using different budget. The performance of the standard cGA is less sensitive to the tested $k$ value on the tested problem.}
\end{subfigure}\hfill
\begin{subfigure}[t]{1\columnwidth}\centering
\includegraphics[width=.8\columnwidth]{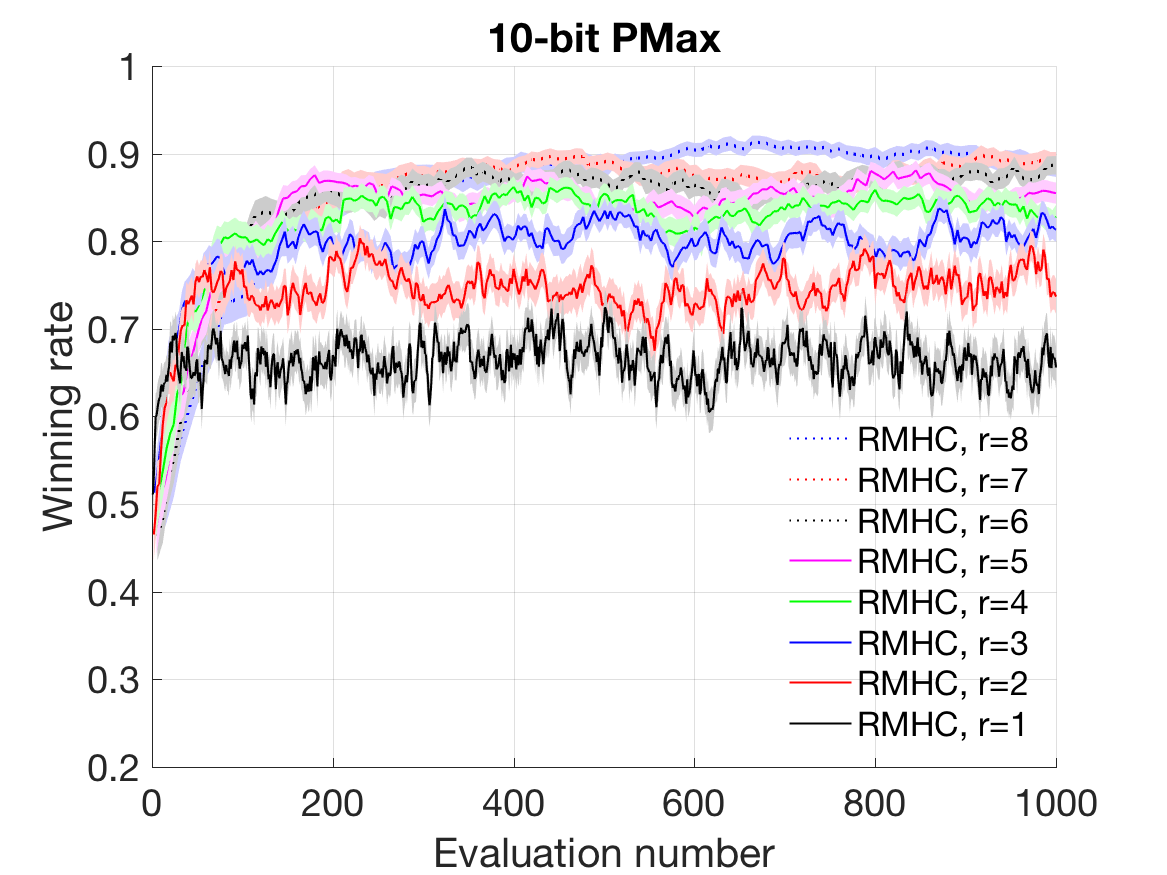}
\caption{Quality of solution recommended by RMHC with resampling number $r$ using different budget. The performance of RMHC without resampling (case $r=1$) is poor as predicted. The optimal resampling number for the tested problem is $4$ for bigger budget and $2$ for smaller budget.}
\end{subfigure}
\caption{\label{fig:baselinepmax}Results of cGA and RMHC on the PMax problem. Each curve is an average of 100 trials. The standard error is also given as a faded area around the average.}
\end{figure*}

The performance of the standard cGA with different virtual population size $k$ and RMHC with different resampling number $r$ on the PMax problem is illustrated in Fig. \ref{fig:baselinepmax}. As the RMHC with variant resampling numbers does not outperform the standard cGA with best setting $k=5d$ (pink curve in Fig. \ref{fig:baselinepmaxcga}), our algorithms are directly compared to the standard cGA with $k=5d$.

\begin{figure*}[htbp]
\centering
 \includegraphics[width=.32\linewidth]{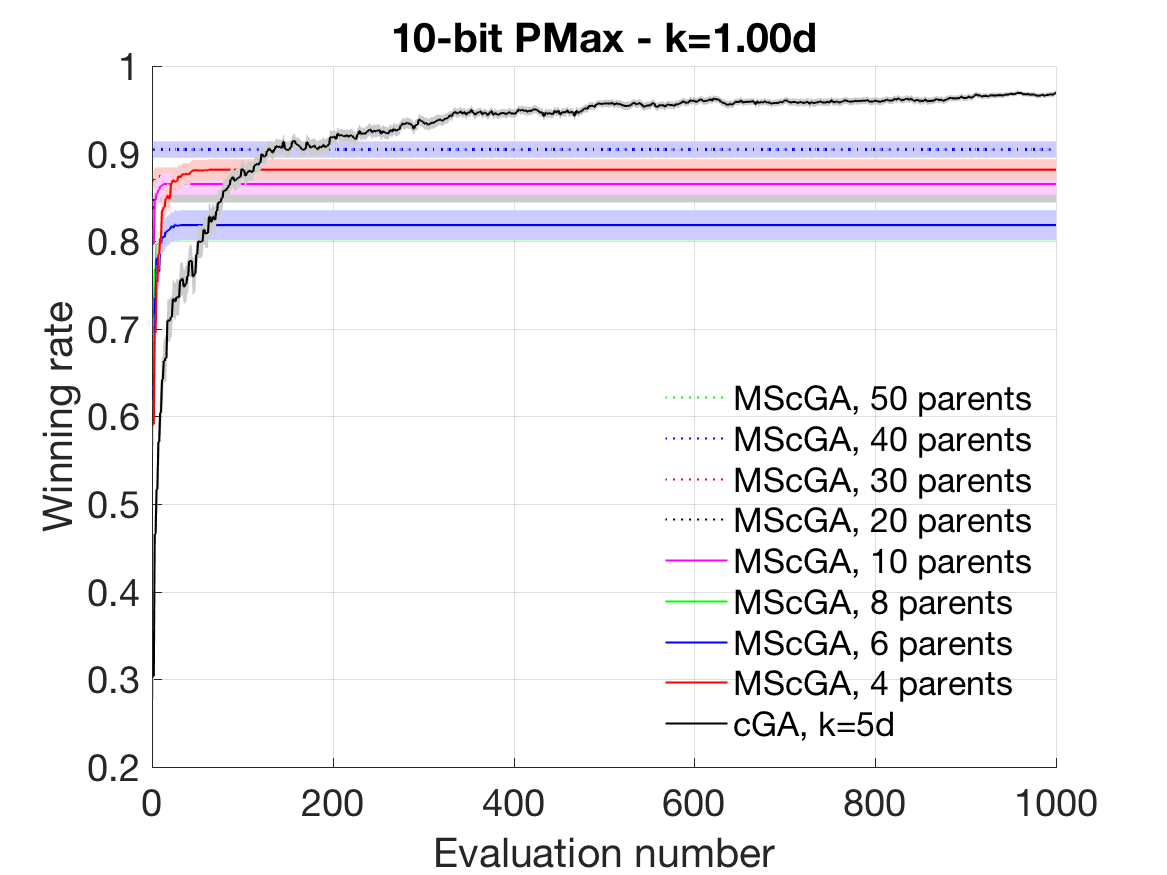}
    \includegraphics[width=.32\linewidth]{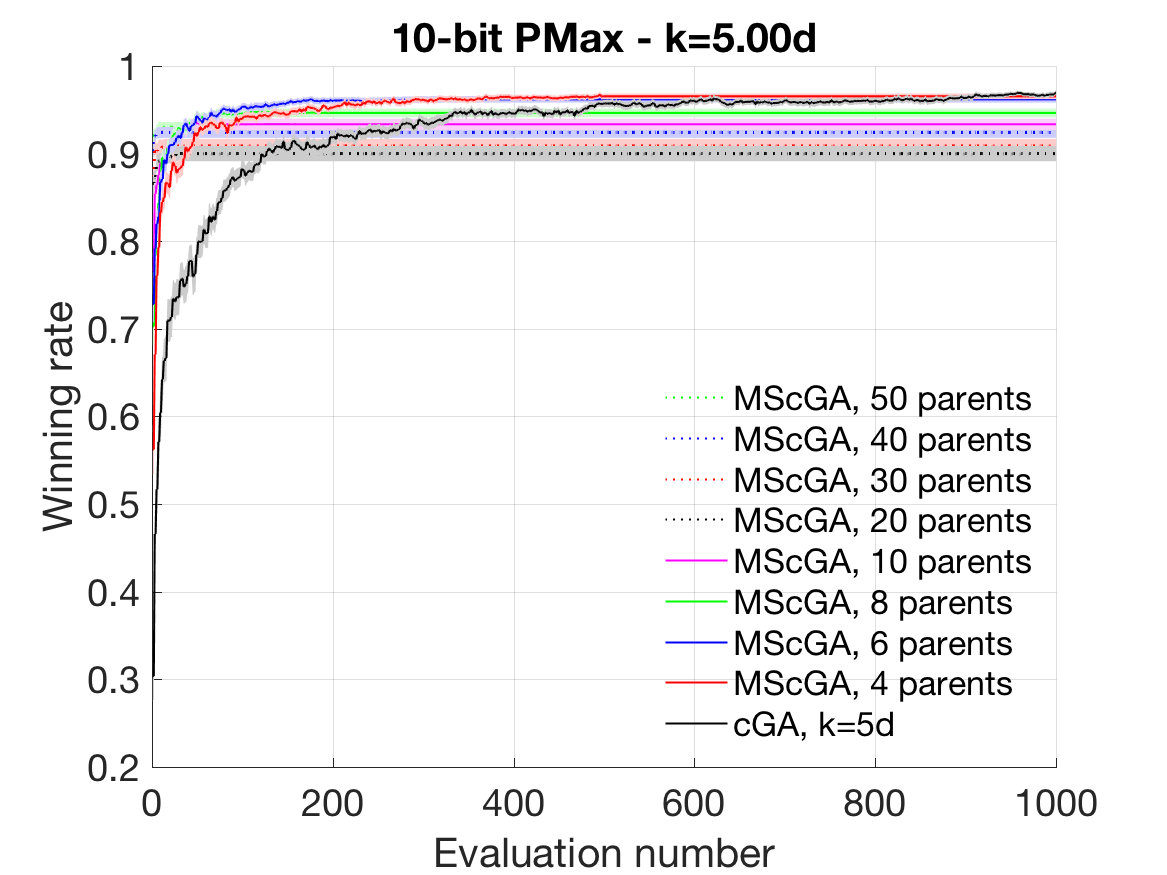}
\includegraphics[width=.32\linewidth]{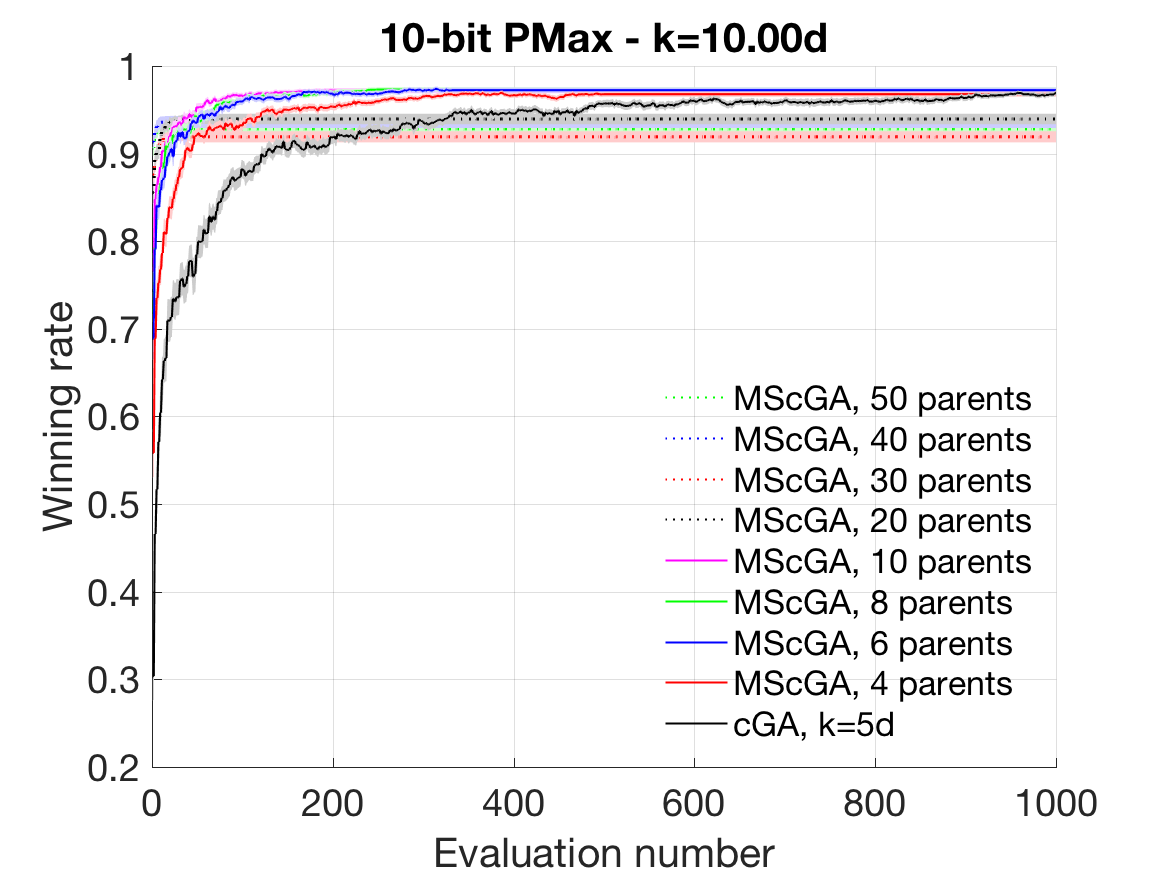}\\
\includegraphics[width=.32\linewidth]{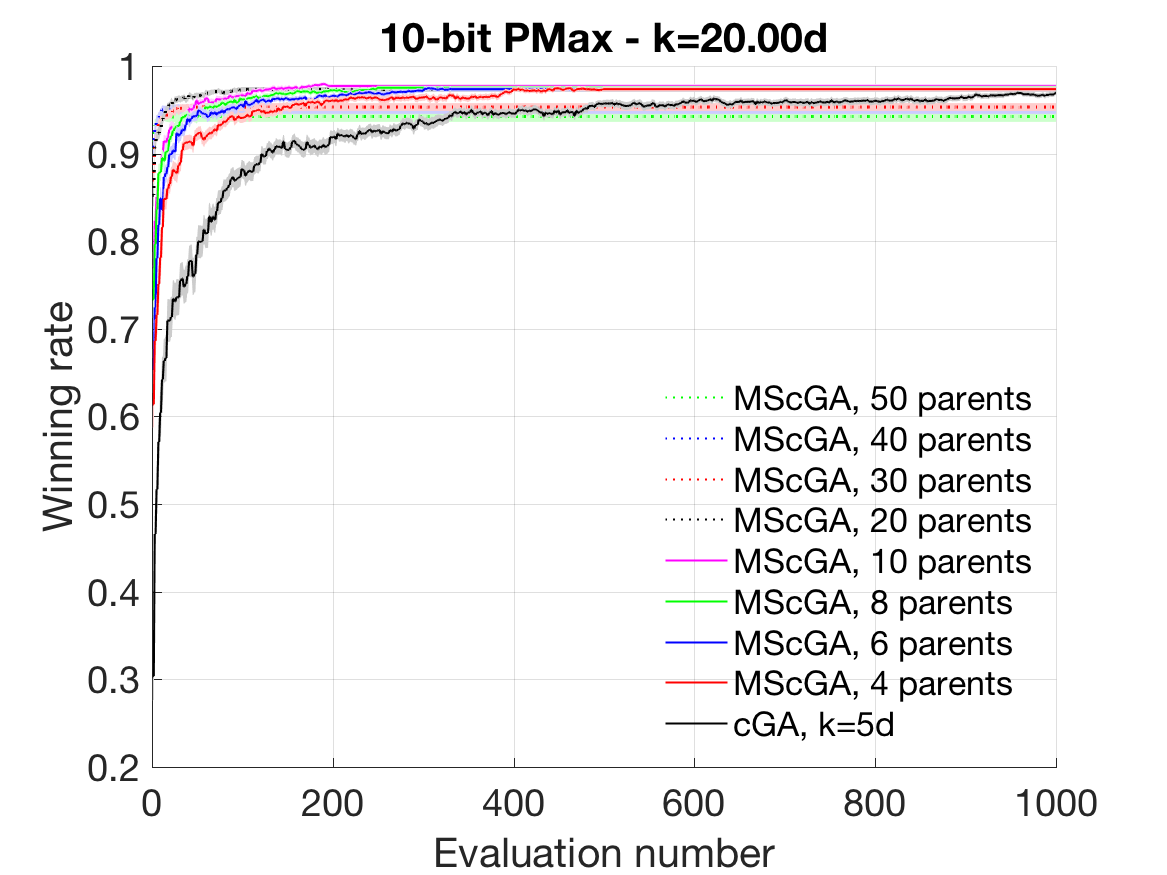}
\includegraphics[width=.32\linewidth]{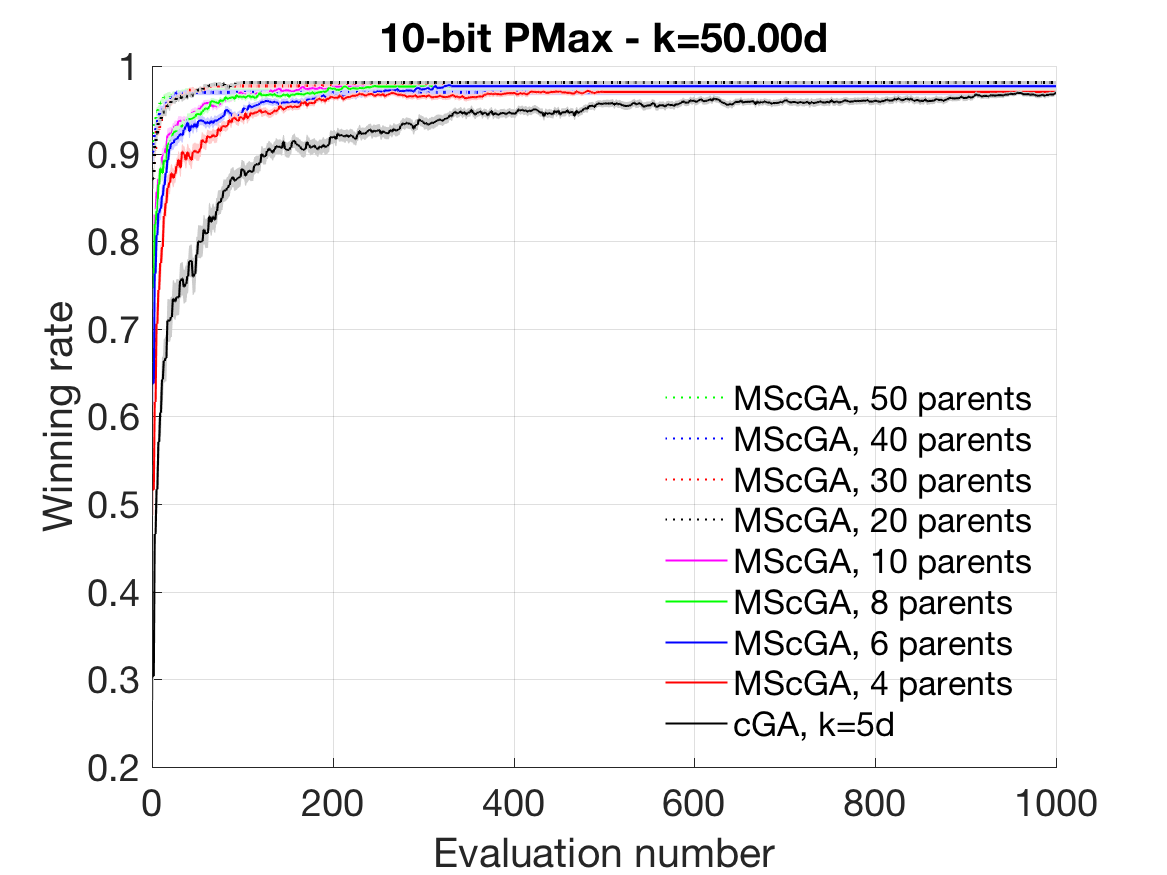}
    \includegraphics[width=.32\linewidth]{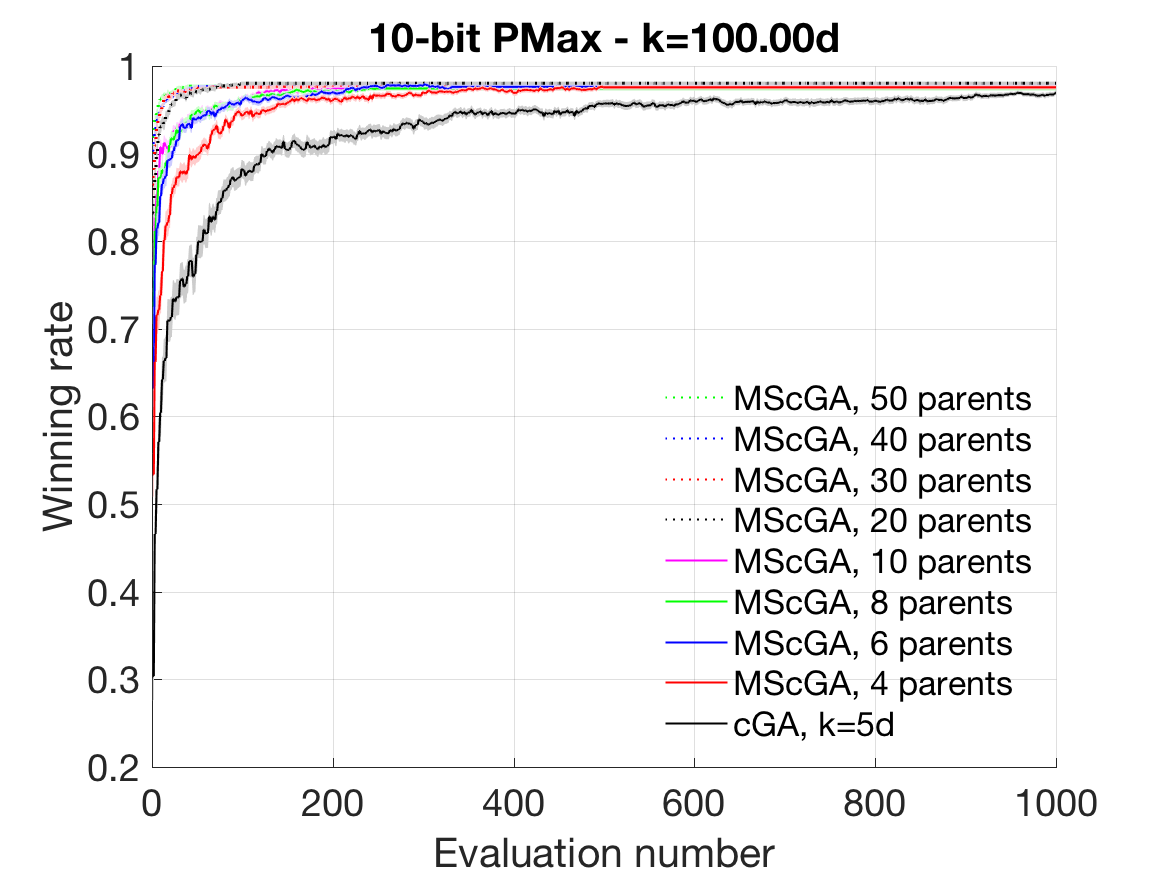}\\
\caption{\label{fig:mscgapmax}Results of the PMax problem optimised by MScGA. Each curve is an average of 100 trials. The standard error is also given as a faded area around the average. The difference of performance of MScGA using different parameter settings is tiny, and all MScGA instances significantly outperform the standard cGA with $k=5d$. The results using other values of $k$ are not shown as they are similar or worse than the one shown.}
\end{figure*}

\begin{figure*}[htbp]
\centering
\includegraphics[width=.32\linewidth]{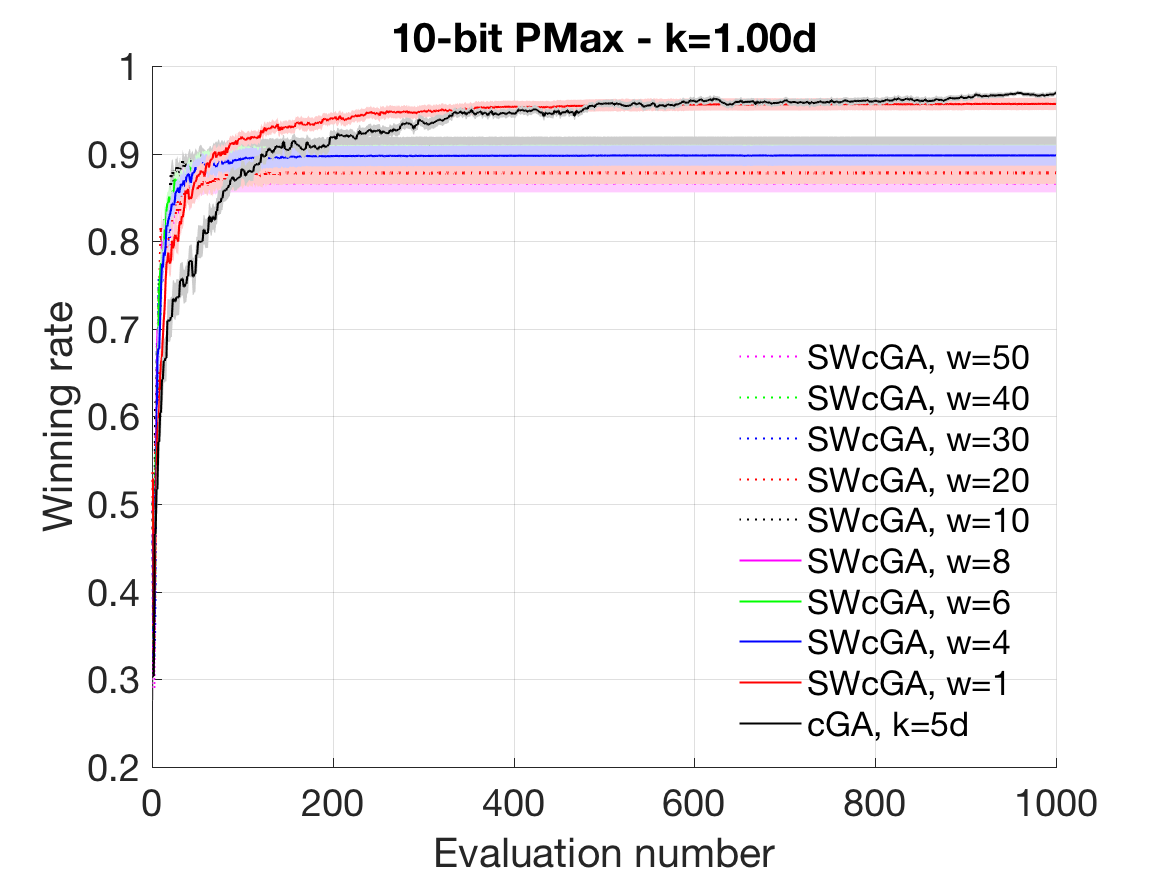}
\includegraphics[width=.32\linewidth]{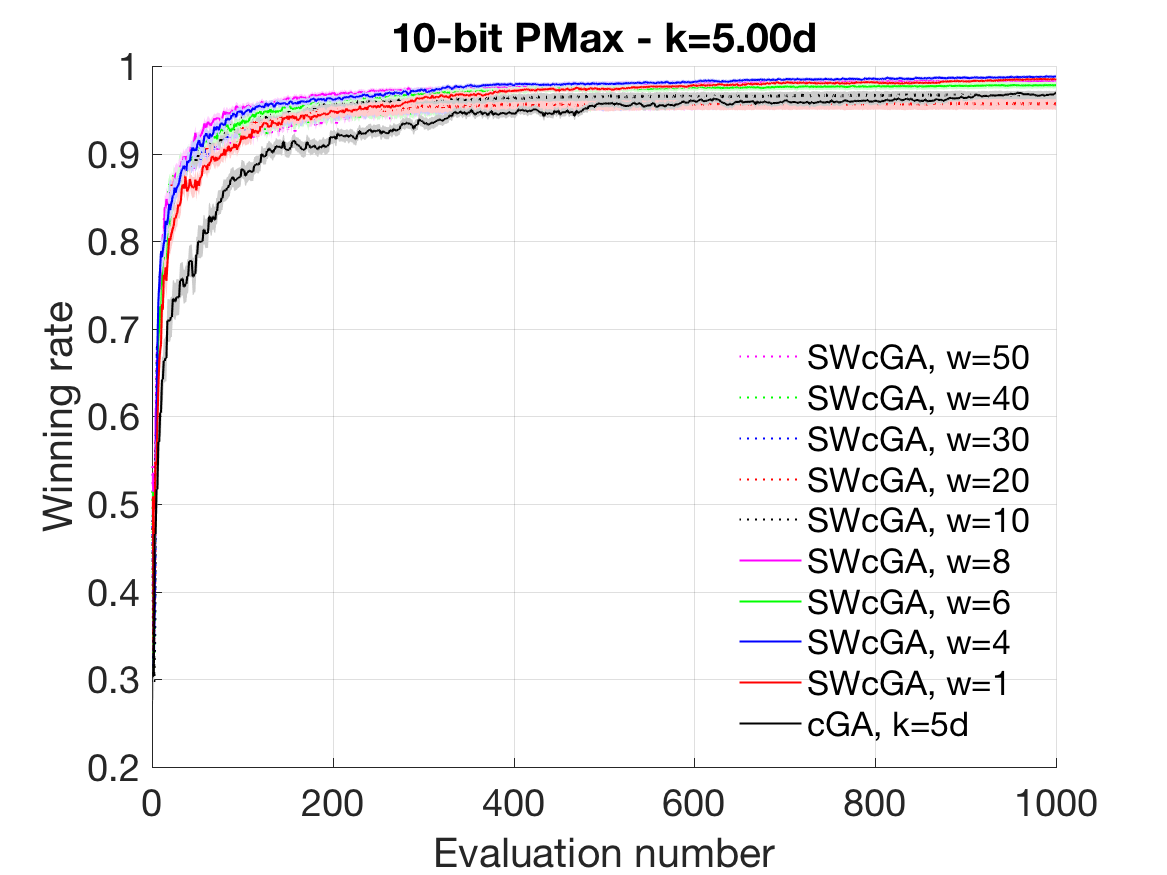}
\includegraphics[width=.32\linewidth]{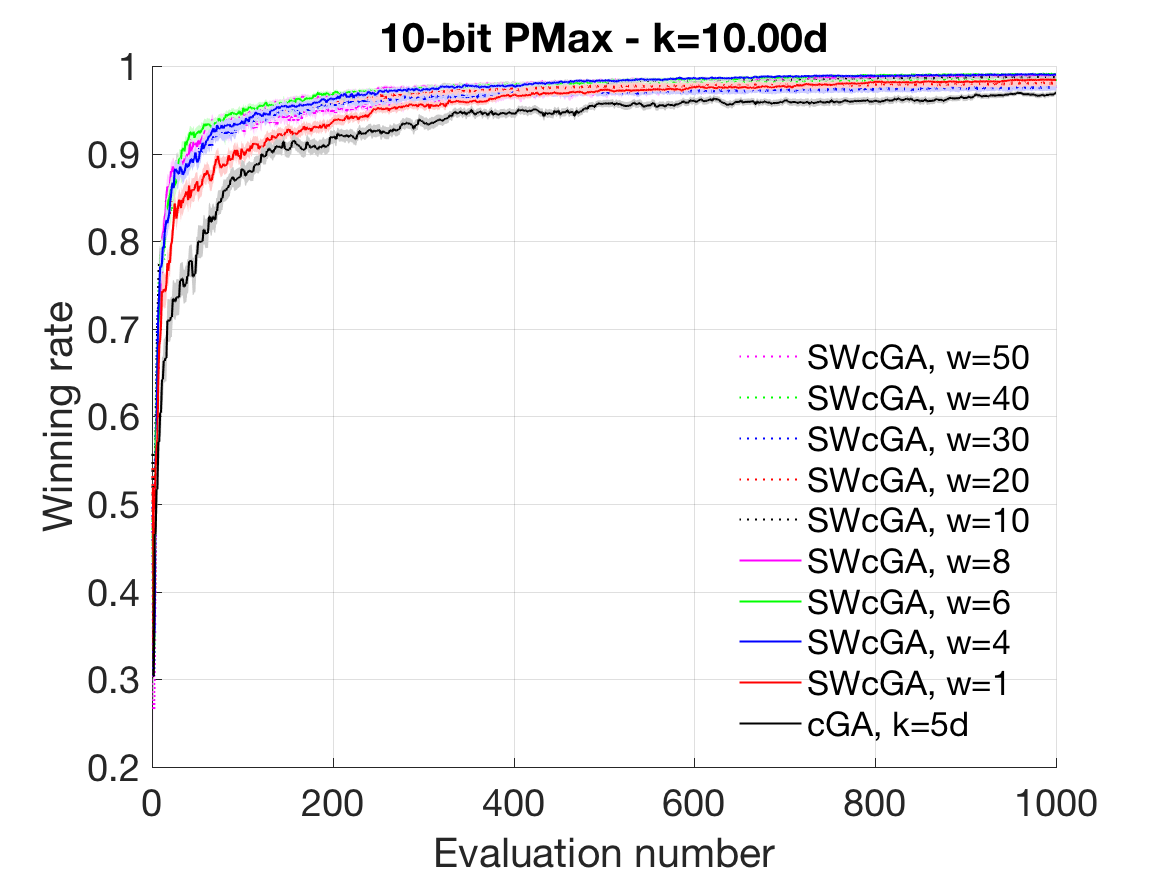}\\
\includegraphics[width=.32\linewidth]{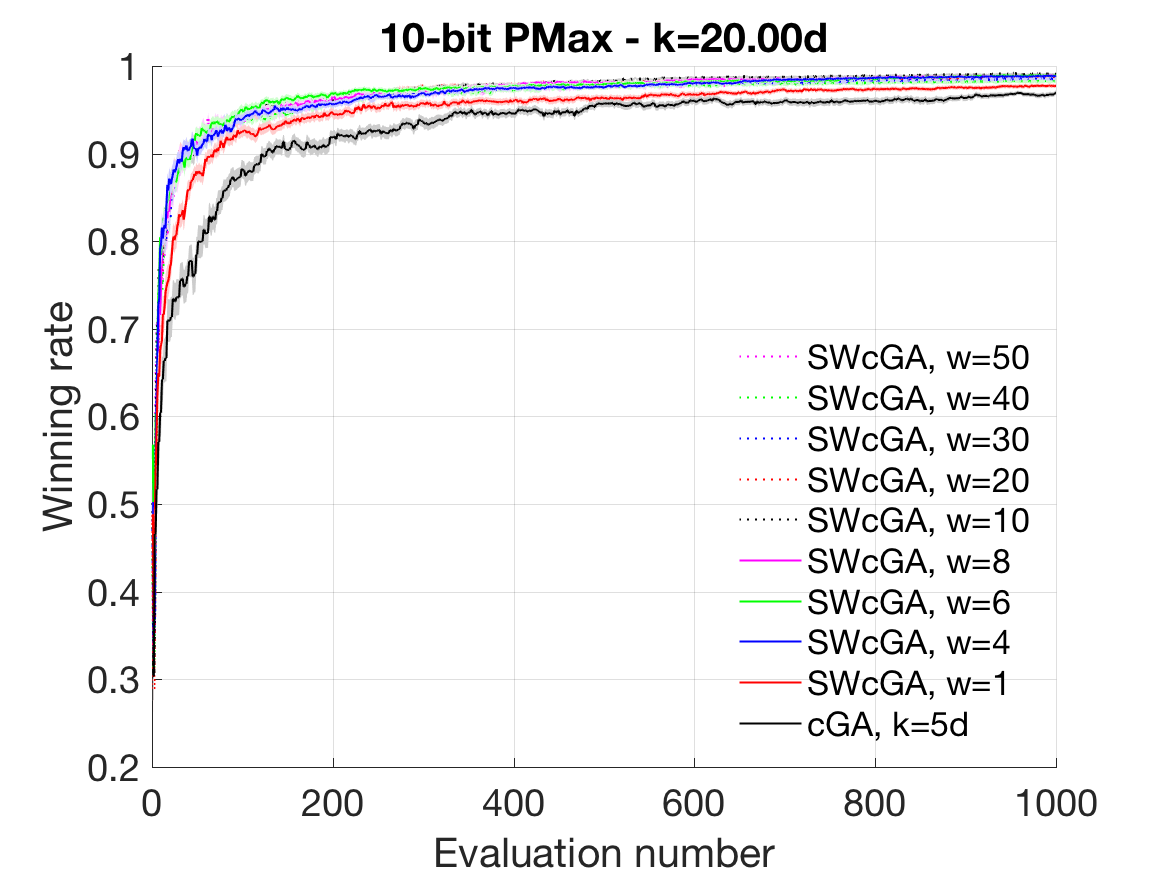}
\includegraphics[width=.32\linewidth]{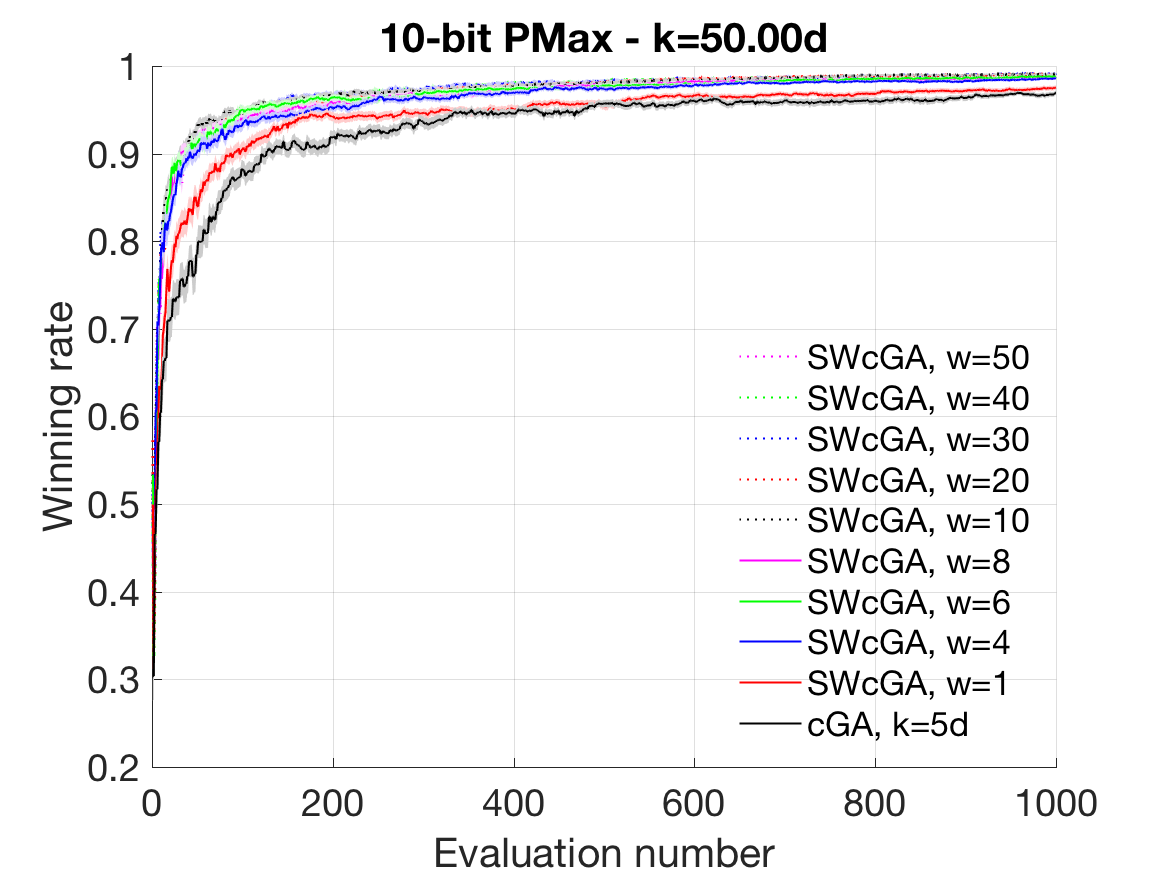}
\includegraphics[width=.32\linewidth]{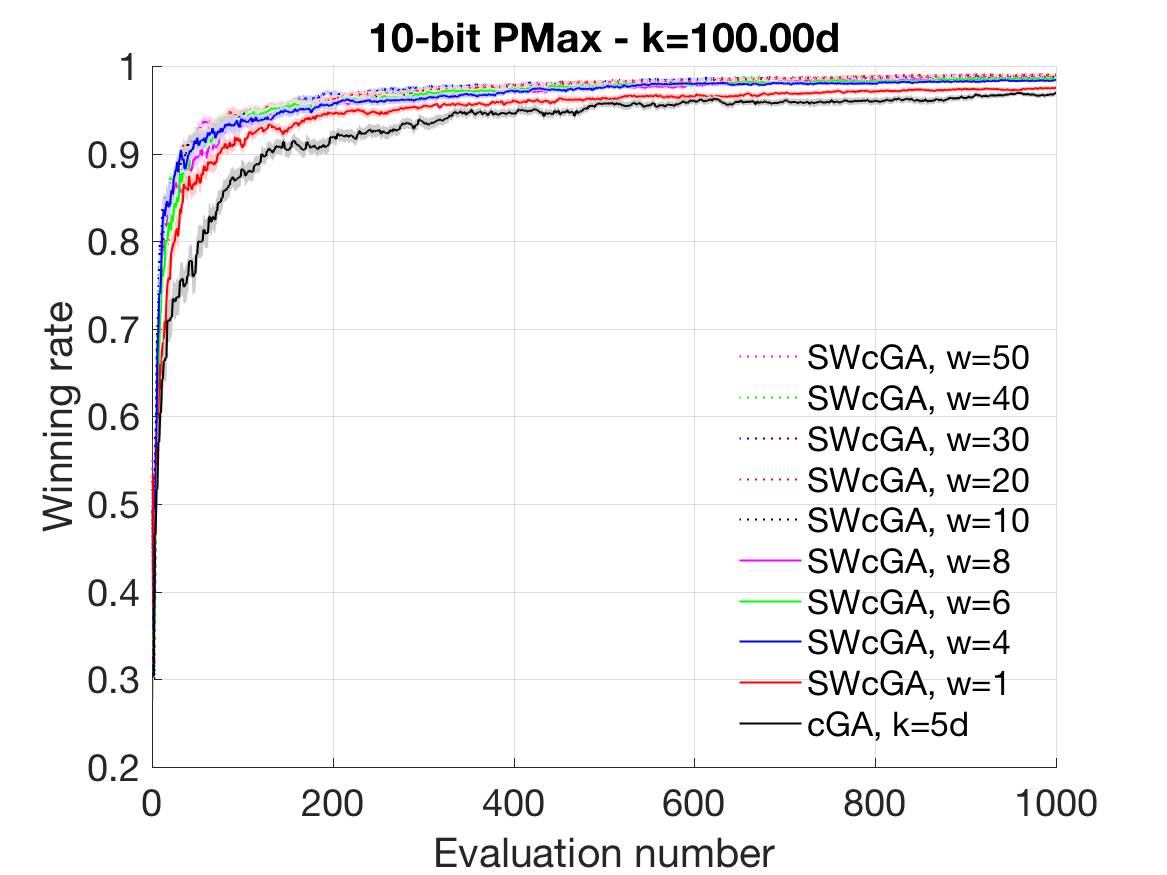}\\
\caption{\label{fig:swcgapmax}Results of the PMax problem optimised by SWcGA. Each curve is an average of 100 trials. The standard error is also given as a faded area around the average. The difference of performance of SWcGA using different parameter settings is tiny, and all SWcGA instances significantly outperform the standard cGA with $k=5d$. The results using other values of $k$ are not shown as they are similar or worse than the one shown.}
\end{figure*}

\begin{figure*}[htbp]
\centering
\begin{subfigure}[t]{1\columnwidth}\centering
\includegraphics[width=.8\columnwidth]{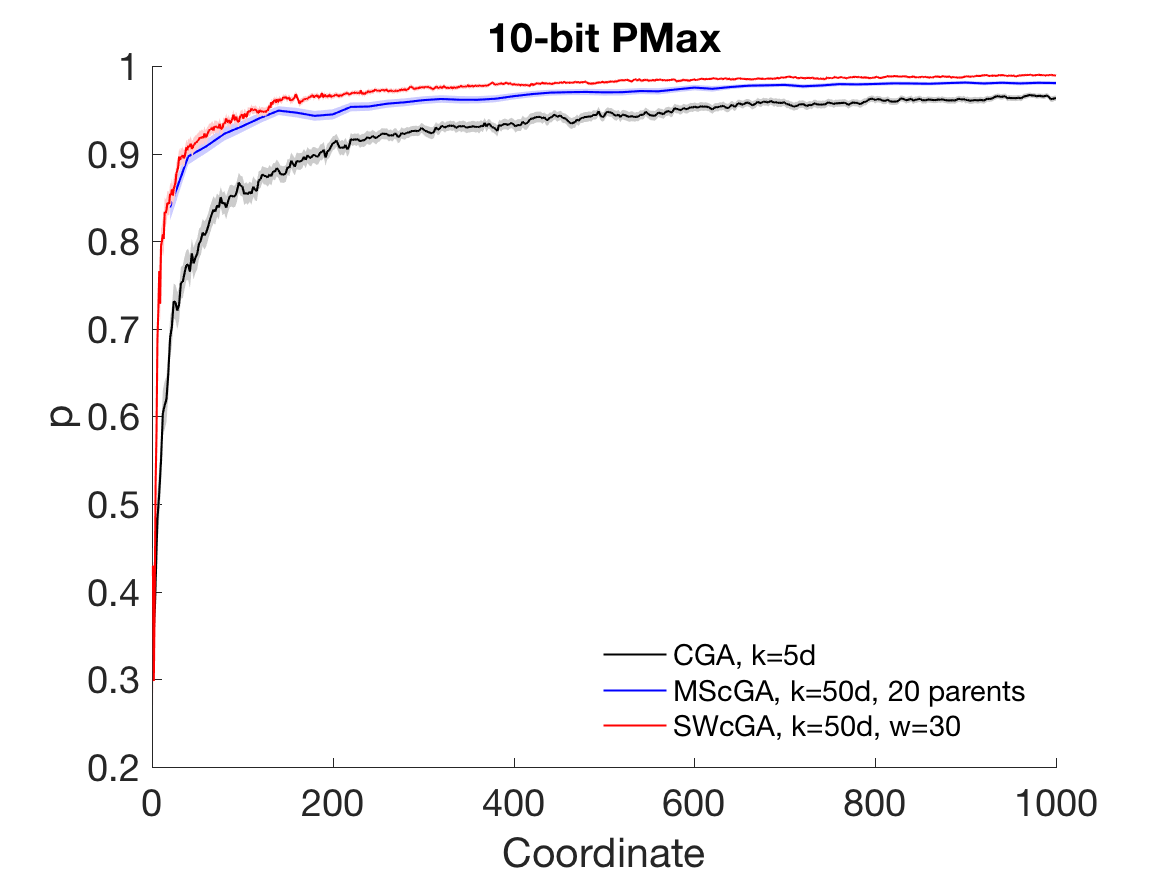}
\caption{Quality of recommendations.}
\end{subfigure}\hfill
\begin{subfigure}[t]{1\columnwidth}\centering
\includegraphics[width=.8\columnwidth]{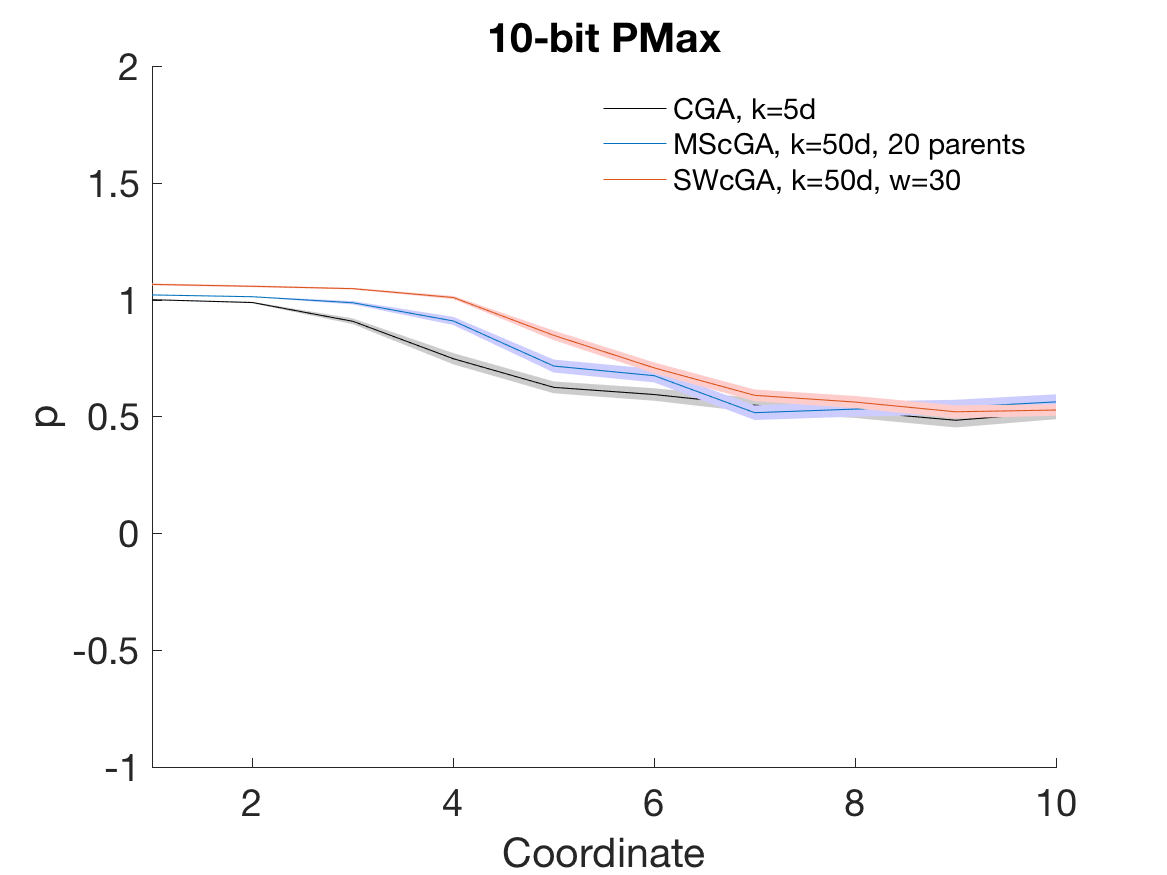}
\caption{Averaged final probability vector $p$.}
\end{subfigure}
\caption{\label{fig:bestpmax}Results of the PMax problem optimised by the cGA, MScGA and WScGA with best tested parameter settings. Each curve is an average of 100 trials. The standard error is also given as a faded area around the average.}
\end{figure*}

The performance of MScGA and SWcGA with different parameter settings are compared to cGA with $k=5d$ (black curves) in Figs. \ref{fig:mscgapmax} and \ref{fig:swcgapmax}, respectively. The best parameter settings of each of the algorithms are listed and compared in Fig. \ref{fig:bestpmax}, as well as the averaged final probability vector $p$ over 100 trials. The SWcGA slightly outperforms the MScGA.

\section{Conclusion and Future Work}
\label{sec:conc}
This paper introduced a simple but important principle to
improve the performance of the compact Genetic Algorithm: to make best possible use of each fitness evaluation by reusing the result in multiple comparisons,
and hence in multiple updates of the probability distribution.

This principle was used to develop two variations of the algorithm:
the first made multiple samples, comparisons and updates at each iteration,
while the second one made just one sample at each iteration, but then performed multiple comparisons and updates by accessing a sliding window of previously evaluated candidates (samples).

Both algorithms significantly outperformed the standard cGA, with the sliding window version performing best. 
The sliding window version is therefore the one we are focusing on in on-going work. 
In addition to offering the best performance at the end of the each run, it also consistently offered better recommendations at nearly
every stage of each run, making it a better choice as an anytime algorithm for use in real-time game AI.  
The sliding window variant is better as an anytime algorithm as it adds only a single candidate
solution per iteration, meaning that the update of the recommendation happens more frequently.

Another interesting observation is the ability of cGA, MScGA and SWcGA to correctly \emph{recommend} the optimal solution without having actually sampled it.  Appendix \ref{app1} summarises the number of times that the optimal solution has been generated among the 100 trials in each experiment. For instance, among the 100 optimisation trials on noisy OneMax by MScGA with $k=20d$ and $n=40$, the optimal solution has been visited only 5 times but the algorithm has never failed in recommending the true optimal solution after $1,000$ fitness evaluations.

We are currently extending the work in two ways.  The first is to allow multi-valued strings, since binary is an unnatural way to represent many problems.  The second is to explore alternative ways to model the probability distribution.  Both of these are already yielding positive results and will be the focus of future research.  Also relevant is our recent work on bandit-based 
optimisation \cite{Kunanusont2017bandit,liu2017bandit}, which explicitly balances exploration versus exploitation, but has not yet been combined with the sliding window approach developed here.  There is reason to believe that such a combination will be beneficial.

\balance
\bibliographystyle{IEEEtran}
\bibliography{main}

\newpage
\appendix[Quality of final recommendation on noisy OneMax problem]\label{app1}
Tables \ref{tab:onemax} and \ref{tab:pmax} summarise the number of times that the optimal solution has been visited over 100 optimisation trials by different algorithms, denoted as ``NHO'' (number of hitting optimum) and the average noise-free fitness of the recommendations after $1000$ evaluations, denoted as ``RQ'' (recommendation quality). It is notable that cGA is the special case of MScGA with $n=2$.

\begin{table}[hbtp]
\centering
\caption{\label{tab:onemax}Noisy OneMax.}
\begin{tabular}{ccccccc}
\hline
\multirow{2}{*}{$k$} & \multicolumn{3}{c}{MCcGA} & \multicolumn{3}{c}{SWcGA}\\
 & $n$ & NHO & RQ & $w$ & NHO & RQ\\
\hline
$50d$ & 50 & 0 & 99.88 $\pm$ 0.04  50 & 6 & 99.99 $\pm$ 0.01 \\
$50d$ & 40 & 0 & 99.91 $\pm$ 0.03  40 & 0 & 99.97 $\pm$ 0.02 \\
$50d$ & 30 & 0 & 99.83 $\pm$ 0.04  30 & 0 & 99.97 $\pm$ 0.02 \\
$50d$ & 20 & 0 & 99.84 $\pm$ 0.04  20 & 0 & 99.83 $\pm$ 0.04 \\
$50d$ & 10 & 0 & 99.64 $\pm$ 0.06  10 & 0 & 99.78 $\pm$ 0.05 \\
$50d$ & 8 & 0 & 99.67 $\pm$ 0.05  8 & 0 & 99.86 $\pm$ 0.04 \\
$50d$ & 6 & 0 & 99.39 $\pm$ 0.08  6 & 0 & 99.67 $\pm$ 0.05 \\
$50d$ & 4 & 0 & 98.94 $\pm$ 0.11  4 & 0 & 99.60 $\pm$ 0.07 \\
$50d$ & 2 & 0 & 95.71 $\pm$ 0.16  1 & 0 & 98.32 $\pm$ 0.11 \\
\hline
$20d$ & 50 & 96 & 99.99 $\pm$ 0.01  50 & 97 & 99.97 $\pm$ 0.02 \\
$20d$ & 40 & 5 & \bf{100.00 $\pm$ 0.00}  40 & 99 & 99.99 $\pm$ 0.01 \\
$20d$ & 30 & 0 & 99.94 $\pm$ 0.03  30 & 99 & 99.99 $\pm$ 0.01 \\
$20d$ & 20 & 0 & 99.92 $\pm$ 0.03  20 & 1 & 99.98 $\pm$ 0.01 \\
$20d$ & 10 & 0 & 99.75 $\pm$ 0.05  10 & 0 & 99.90 $\pm$ 0.03 \\
$20d$ & 8 & 0 & 99.63 $\pm$ 0.06  8 & 0 & 99.81 $\pm$ 0.04 \\
$20d$ & 6 & 0 & 99.48 $\pm$ 0.07  6 & 0 & 99.70 $\pm$ 0.05 \\
$20d$ & 4 & 0 & 99.08 $\pm$ 0.10  4 & 0 & 99.63 $\pm$ 0.06 \\
$20d$ & 2 & 0 & 95.70 $\pm$ 0.19  1 & 0 & 98.40 $\pm$ 0.12 \\
\hline
$10d$ & 50 & 98 & 99.98 $\pm$ 0.01  50 & 0 & 88.64 $\pm$ 0.27 \\
$10d$ & 40 & 100 & \bf{100.00 $\pm$ 0.00}  40 & 25 & 98.43 $\pm$ 0.13 \\
$10d$ & 30 & 100 & \bf{100.00 $\pm$ 0.00}  30 & 88 & 99.86 $\pm$ 0.04 \\
$10d$ & 20 & 2 & 99.99 $\pm$ 0.01  20 & 98 & 99.98 $\pm$ 0.01 \\
$10d$ & 10 & 0 & 99.77 $\pm$ 0.05  10 & 4 & 99.98 $\pm$ 0.01 \\
$10d$ & 8 & 0 & 99.70 $\pm$ 0.05  8 & 0 & 99.95 $\pm$ 0.02 \\
$10d$ & 6 & 0 & 99.43 $\pm$ 0.07  6 & 0 & 99.86 $\pm$ 0.04 \\
$10d$ & 4 & 0 & 99.06 $\pm$ 0.11  4 & 0 & 99.76 $\pm$ 0.06 \\
$10d$ & 2 & 0 & 95.67 $\pm$ 0.20  1 & 0 & 98.22 $\pm$ 0.12 \\
\hline
$5d$ & 50 & 51 & 99.23 $\pm$ 0.10  50 & 0 & 73.95 $\pm$ 0.29 \\
$5d$ & 40 & 73 & 99.70 $\pm$ 0.05  40 & 0 & 79.11 $\pm$ 0.26 \\
$5d$ & 30 & 94 & 99.93 $\pm$ 0.03  30 & 0 & 93.71 $\pm$ 0.22 \\
$5d$ & 20 & 100 & \bf{100.00 $\pm$ 0.00}  20 & 63 & 99.52 $\pm$ 0.07 \\
$5d$ & 10 & 0 & 99.95 $\pm$ 0.02  10 & 100 & \bf{100.00 $\pm$ 0.00} \\
$5d$ & 8 & 0 & 99.84 $\pm$ 0.04  8 & 100 & \bf{100.00 $\pm$ 0.00} \\
$5d$ & 6 & 0 & 99.64 $\pm$ 0.06  6 & 88 & \bf{100.00 $\pm$ 0.00} \\
$5d$ & 4 & 0 & 99.19 $\pm$ 0.10  4 & 0 & 99.95 $\pm$ 0.02 \\
$5d$ & 2 & 0 & 95.69 $\pm$ 0.19  1 & 0 & 98.29 $\pm$ 0.11 \\
\hline
$2d$ & 50 & 0 & 93.11 $\pm$ 0.24  50 & 0 & 68.44 $\pm$ 0.34 \\
$2d$ & 40 & 0 & 94.29 $\pm$ 0.23  40 & 0 & 68.43 $\pm$ 0.30 \\
$2d$ & 30 & 3 & 96.70 $\pm$ 0.17  30 & 0 & 71.29 $\pm$ 0.28 \\
$2d$ & 20 & 40 & 99.06 $\pm$ 0.10  20 & 0 & 84.46 $\pm$ 0.28 \\
$2d$ & 10 & 98 & 99.97 $\pm$ 0.02  10 & 29 & 98.75 $\pm$ 0.10 \\
$2d$ & 8 & 100 & 99.99 $\pm$ 0.01  8 & 61 & 99.48 $\pm$ 0.07 \\
$2d$ & 6 & 96 & 99.99 $\pm$ 0.01  6 & 84 & 99.83 $\pm$ 0.04 \\
$2d$ & 4 & 0 & 99.58 $\pm$ 0.06  4 & 99 & 99.99 $\pm$ 0.01 \\
$2d$ & 2 & 0 & 96.12 $\pm$ 0.18  1 & 0 & 98.72 $\pm$ 0.09 \\
\hline
$d$ & 50 & 0 & 85.81 $\pm$ 0.33  50 & 0 & 65.28 $\pm$ 0.31 \\
$d$ & 40 & 0 & 86.76 $\pm$ 0.29  40 & 0 & 66.03 $\pm$ 0.33 \\
$d$ & 30 & 0 & 89.25 $\pm$ 0.27  30 & 0 & 66.48 $\pm$ 0.34 \\
$d$ & 20 & 0 & 93.58 $\pm$ 0.24  20 & 0 & 69.45 $\pm$ 0.27 \\
$d$ & 20 & 30 & 98.81 $\pm$ 0.11  10 & 0 & 91.44 $\pm$ 0.24 \\
$d$ & 8 & 58 & 99.50 $\pm$ 0.06  8 & 0 & 94.75 $\pm$ 0.21 \\
$d$ & 6 & 87 & 99.86 $\pm$ 0.04  6 & 5 & 97.15 $\pm$ 0.14 \\
$d$ & 4 & 99 & 99.98 $\pm$ 0.01  4 & 41 & 99.18 $\pm$ 0.08 \\
$d$ & 2 & 0 & 96.73 $\pm$ 0.16  1 & 2 & 99.61 $\pm$ 0.06 \\
\hline
$d/2$ & 50 & 0 & 81.00 $\pm$ 0.32  50 & 0 & 63.94 $\pm$ 0.32 \\
$d/2$ & 40 & 0 & 81.69 $\pm$ 0.32  40 & 0 & 63.58 $\pm$ 0.29 \\
$d/2$ & 30 & 0 & 81.41 $\pm$ 0.32  30 & 0 & 64.05 $\pm$ 0.30 \\
$d/2$ & 20 & 0 & 84.25 $\pm$ 0.30  20 & 0 & 64.10 $\pm$ 0.28 \\
$d/2$ & 10 & 0 & 92.35 $\pm$ 0.22  10 & 0 & 74.91 $\pm$ 0.32 \\
$d/2$ & 8 & 0 & 94.47 $\pm$ 0.19  8 & 0 & 82.87 $\pm$ 0.31 \\
$d/2$ & 6 & 4 & 96.97 $\pm$ 0.15  6 & 0 & 88.10 $\pm$ 0.26 \\
$d/2$ & 4 & 39 & 99.01 $\pm$ 0.10  4 & 0 & 93.42 $\pm$ 0.21 \\
$d/2$ & 2 & 0 & \bf{98.68 $\pm$ 0.12}  1 & 72 & 99.68 $\pm$ 0.05 \\
\hline
\end{tabular}
\end{table}

\begin{table}[hbtp]
\centering
\caption{\label{tab:pmax}PMax.}
\begin{tabular}{ccccccc}
\hline
\multirow{2}{*}{$k$} & \multicolumn{3}{c}{MCcGA} & \multicolumn{3}{c}{SWcGA}\\
 & $n$ & NHO & RQ & $w$ & NHO & RQ\\
\hline
$100d$ & 50 & 88 & 0.9770 $\pm$ 0.0020 & 50 & 97 & 0.9880 $\pm$ 0.0012 \\
$100d$ & 40 & 92 & 0.9765 $\pm$ 0.0021 & 40 & 97 & 0.9897 $\pm$ 0.0010 \\
$100d$ & 30 & 91 & 0.9760 $\pm$ 0.0020 & 30 & 98 & 0.9881 $\pm$ 0.0012 \\
$100d$ & 20 & 98 & 0.9803 $\pm$ 0.0020 & 20 & 99 & 0.9899 $\pm$ 0.0010 \\
$100d$ & 10 & 97 & 0.9741 $\pm$ 0.0023 & 10 & 98 & 0.9870 $\pm$ 0.0013 \\
$100d$ & 8 & 94 & 0.9745 $\pm$ 0.0025  & 8 & 99 & 0.9857 $\pm$ 0.0014 \\
$100d$ & 6 & 88 & 0.9762 $\pm$ 0.0025  & 6 & 100 & 0.9853 $\pm$ 0.0015 \\
$100d$ & 4 & 76 & 0.9758 $\pm$ 0.0023  & 4 & 99 & 0.9846 $\pm$ 0.0015 \\
$100d$ & 2 & 67 & 0.9680 $\pm$ 0.0026  & 1 & 83 & 0.9755 $\pm$ 0.0022 \\
\hline
$50d$ & 50 & 57 & 0.9713 $\pm$ 0.0033 & 50 & 93 & 0.9875 $\pm$ 0.0013 \\
$50d$ & 40 & 66 & 0.9700 $\pm$ 0.0032 & 40 & 93 & 0.9884 $\pm$ 0.0011 \\
$50d$ & 30 & 78 & 0.9773 $\pm$ 0.0021 & 30 & 94 & \bf{0.9919 $\pm$ 0.0009} \\
$50d$ & 20 & 95 & \bf{0.9812 $\pm$ 0.0018 } & 20 & 97 & 0.9902 $\pm$ 0.0010 \\
$50d$ & 10 & 97 & 0.9763 $\pm$ 0.0021 & 10 & 99 & 0.9908 $\pm$ 0.0008 \\
$50d$ & 8 & 100 & 0.9766 $\pm$ 0.0020 & 8 & 100 & 0.9876 $\pm$ 0.0013 \\
$50d$ & 6 & 100 & 0.9770 $\pm$ 0.0024 & 6 & 99 & 0.9882 $\pm$ 0.0012 \\
$50d$ & 4 & 94 & 0.9704 $\pm$ 0.0026 & 4 & 100 & 0.9857 $\pm$ 0.0015 \\
$50d$ & 2 & 69 & 0.9682 $\pm$ 0.0028 & 1 & 85 & 0.9753 $\pm$ 0.0026 \\
\hline
$20d$ & 50 & 21 & 0.9425 $\pm$ 0.0061 & 50 & 68 & 0.9835 $\pm$ 0.0020 \\
$20d$ & 40 & 23 & 0.9507 $\pm$ 0.0053 & 40 & 65 & 0.9832 $\pm$ 0.0027 \\
$20d$ & 30 & 27 & 0.9532 $\pm$ 0.0044 & 30 & 68 & 0.9861 $\pm$ 0.0021 \\
$20d$ & 20 & 53 & 0.9733 $\pm$ 0.0028 & 20 & 68 & 0.9882 $\pm$ 0.0014 \\
$20d$ & 10 & 87 & 0.9772 $\pm$ 0.0022 & 10 & 90 & 0.9908 $\pm$ 0.0011 \\
$20d$ & 8 & 86 & 0.9753 $\pm$ 0.0023 & 8 & 93 & 0.9887 $\pm$ 0.0012 \\
$20d$ & 6 & 94 & 0.9736 $\pm$ 0.0024 & 6 & 94 & 0.9889 $\pm$ 0.0010 \\
$20d$ & 4 & 99 & 0.9737 $\pm$ 0.0025 & 4 & 99 & 0.9887 $\pm$ 0.0011 \\
$20d$ & 2 & 89 & 0.9621 $\pm$ 0.0038 & 1 & 98 & 0.9774 $\pm$ 0.0023 \\
\hline
$10d$ & 50 & 14 & 0.9283 $\pm$ 0.0068 & 50 & 48 & 0.9769 $\pm$ 0.0029 \\
$10d$ & 40 & 9 & 0.9362 $\pm$ 0.0068 & 40 & 62 & 0.9828 $\pm$ 0.0026 \\
$10d$ & 30 & 13 & 0.9192 $\pm$ 0.0063 & 30 & 47 & 0.9759 $\pm$ 0.0038 \\
$10d$ & 20 & 13 & 0.9396 $\pm$ 0.0059 & 20 & 42 & 0.9801 $\pm$ 0.0038 \\
$10d$ & 10 & 53 & 0.9719 $\pm$ 0.0030 & 10 & 65 & 0.9878 $\pm$ 0.0023 \\
$10d$ & 8 & 53 & 0.9733 $\pm$ 0.0025 & 8 & 68 & 0.9899 $\pm$ 0.0013 \\
$10d$ & 6 & 76 & 0.9725 $\pm$ 0.0028 & 6 & 83 & 0.9909 $\pm$ 0.0011 \\
$10d$ & 4 & 93 & 0.9679 $\pm$ 0.0029 & 4 & 92 & 0.9905 $\pm$ 0.0012 \\
$10d$ & 2 & 98 & 0.9685 $\pm$ 0.0033 & 1 & 97 & 0.9841 $\pm$ 0.0016 \\
\hline
$5d$ & 50 & 5 & 0.9301 $\pm$ 0.0061 & 50 & 31 & 0.9559 $\pm$ 0.0045 \\
$5d$ & 40 & 11 & 0.9243 $\pm$ 0.0063 & 40 & 34 & 0.9578 $\pm$ 0.0053 \\
$5d$ & 30 & 5 & 0.9088 $\pm$ 0.0081 & 30 & 42 & 0.9585 $\pm$ 0.0066 \\
$5d$ & 20 & 9 & 0.9001 $\pm$ 0.0087 & 20 & 35 & 0.9570 $\pm$ 0.0069 \\
$5d$ & 10 & 23 & 0.9336 $\pm$ 0.0064 & 10 & 26 & 0.9669 $\pm$ 0.0046 \\
$5d$ & 8 & 40 & 0.9465 $\pm$ 0.0049 & 8 & 42 & 0.9831 $\pm$ 0.0026 \\
$5d$ & 6 & 36 & 0.9621 $\pm$ 0.0037 & 6 & 42 & 0.9783 $\pm$ 0.0037 \\
$5d$ & 4 & 55 & 0.9655 $\pm$ 0.0032 & 4 & 68 & 0.9881 $\pm$ 0.0019 \\
$5d$ & 2 & 89 & 0.9694 $\pm$ 0.0025 & 1 & 96 & 0.9844 $\pm$ 0.0015 \\
\hline
$2d$ & 50 & 4 & 0.8974 $\pm$ 0.0096 & 50 & 12 & 0.9128 $\pm$ 0.0078 \\
$2d$ & 40 & 2 & 0.9041 $\pm$ 0.0086 & 40 & 18 & 0.9294 $\pm$ 0.0068 \\
$2d$ & 30 & 6 & 0.8763 $\pm$ 0.0097 & 30 & 20 & 0.9268 $\pm$ 0.0093 \\
$2d$ & 20 & 4 & 0.8926 $\pm$ 0.0102 & 20 & 10 & 0.9173 $\pm$ 0.0100 \\
$2d$ & 10 & 7 & 0.8770 $\pm$ 0.0117 & 10 & 12 & 0.9413 $\pm$ 0.0074 \\
$2d$ & 8 & 11 & 0.8788 $\pm$ 0.0125 & 8 & 13 & 0.9418 $\pm$ 0.0073 \\
$2d$ & 6 & 16 & 0.8896 $\pm$ 0.0114 & 6 & 12 & 0.9376 $\pm$ 0.0084 \\
$2d$ & 4 & 22 & 0.9223 $\pm$ 0.0082 & 4 & 25 & 0.9585 $\pm$ 0.0058 \\
$2d$ & 2 & 38 & 0.9528 $\pm$ 0.0046 & 1 & 57 & 0.9787 $\pm$ 0.0032 \\
\hline
$d$ & 50 & 5 & 0.9043 $\pm$ 0.0091 & 50 & 3 & 0.8664 $\pm$ 0.0106 \\
$d$ & 40 & 2 & 0.9048 $\pm$ 0.0093 & 40 & 3 & 0.8812 $\pm$ 0.0122 \\
$d$ & 30 & 3 & 0.8739 $\pm$ 0.0104 & 30 & 6 & 0.8784 $\pm$ 0.0117 \\
$d$ & 20 & 3 & 0.8567 $\pm$ 0.0125 & 20 & 5 & 0.8780 $\pm$ 0.0126 \\
$d$ & 10 & 3 & 0.8654 $\pm$ 0.0123 & 10 & 8 & 0.9082 $\pm$ 0.0114 \\
$d$ & 8 & 0 & 0.8144 $\pm$ 0.0140 & 8 & 10 & 0.8988 $\pm$ 0.0111 \\
$d$ & 6 & 5 & 0.8185 $\pm$ 0.0169 & 6 & 2 & 0.9002 $\pm$ 0.0100 \\
$d$ & 4 & 10 & 0.8817 $\pm$ 0.0120 & 4 & 5 & 0.8980 $\pm$ 0.0116 \\
$d$ & 2 & 23 & 0.9140 $\pm$ 0.0083 & 1 & 20 & 0.9568 $\pm$ 0.0070 \\
\hline
$d/2$ & 50 & 5 & 0.9113 $\pm$ 0.0087 & 50 & 4 & 0.8349 $\pm$ 0.0148 \\
$d/2$ & 40 & 1 & 0.8876 $\pm$ 0.0087 & 40 & 4 & 0.8288 $\pm$ 0.0156 \\
$d/2$ & 30 & 0 & 0.8778 $\pm$ 0.0107 & 30 & 2 & 0.8368 $\pm$ 0.0148 \\
$d/2$ & 20 & 4 & 0.8417 $\pm$ 0.0164 & 20 & 6 & 0.8304 $\pm$ 0.0161 \\
$d/2$ & 10 & 3 & 0.8113 $\pm$ 0.0162 & 10 & 2 & 0.8435 $\pm$ 0.0158 \\
$d/2$ & 8 & 3 & 0.8033 $\pm$ 0.0197 & 8 & 6 & 0.8602 $\pm$ 0.0147 \\
$d/2$ & 6 & 1 & 0.8100 $\pm$ 0.0180 & 6 & 5 & 0.8555 $\pm$ 0.0139 \\
$d/2$ & 4 & 2 & 0.7684 $\pm$ 0.0220 & 4 & 4 & 0.8676 $\pm$ 0.0123 \\
$d/2$ & 2 & 16 & 0.8493 $\pm$ 0.0152 & 1 & 12 & 0.9021 $\pm$ 0.0123 \\
\hline
\end{tabular}
\end{table}

\end{document}